%% file: main.tex
\documentclass[10pt,twocolumn,letterpaper]{article}

\input{setup/package}
\input{setup/macros}

\input{setup/symbols}

\input{setup/graphicspath}

\makeatletter

\def\@fnsymbol#1{\ensuremath{\ifcase#1\or \dagger\or \ddagger\or
\mathsection\or \mathparagraph\or \|\or **\or \dagger\dagger
\or \ddagger\ddagger \else\@ctrerr\fi}}
\makeatother

\graphicspath{{figures/}}

\usepackage[accsupp]{axessibility}  
\usepackage[pagenumbers]{cvpr} 

\def\cvprPaperID{2831} 
\def\confName{CVPR}
\def\confYear{2023}

\begin{document}

\title{Progressively Optimized Local Radiance Fields for Robust View Synthesis}

\author{
Andreas Meuleman$^{1}$\footnotemark[2]\quad 
Yu-Lun Liu$^{2}$\addtocounter{footnote}{-1}\addtocounter{Hfootnote}{-1}\footnotemark[2]\quad 
Chen Gao$^{3}$\\
Jia-Bin Huang$^{3,4}$\quad 
Changil Kim$^{3}$\quad 
Min H. Kim$^{1}$\quad 
Johannes Kopf$^{3}$\\
$^{1}$KAIST \quad
$^{2}$National Taiwan University \quad
$^{3}$Meta \quad
$^{4}$University of Maryland, College Park\\
{\url{https://localrf.github.io/}}
}

\twocolumn[{
\renewcommand\twocolumn[1][]{#1}
\maketitle
\input{figure/fig_teaser}
}]
\renewcommand{\thefootnote}{\fnsymbol{footnote}}
\footnotetext[2]{Part of the work was done while Andreas and Yu-Lun were interns at Meta.}

\maketitle

\thispagestyle{empty}
\input{0_abstract}

\input{1_introduction}
\input{2_related}
\input{3_method}
\input{4_result}
\input{5_limitations}
\input{6_conclusions}

\clearpage

{\small
\bibliographystyle{ieee_fullname}
\bibliography{egbib}
}

\end{document}

%% file: setup/package.tex
\usepackage{everyshi}

\usepackage{dsfont}
\usepackage{graphicx}
\usepackage{subcaption}
\usepackage{float}
\usepackage{lscape}                                         
\usepackage{wrapfig}
\usepackage[percent]{overpic}
\usepackage{tikz}

\usepackage[lined,ruled,linesnumbered]{algorithm2e}
\usepackage{animate} 

\usepackage{booktabs}                   
\usepackage{multirow}

\usepackage{makecell}

\usepackage{paralist}
\usepackage{enumitem}
\setlist[enumerate]{itemsep=0mm}

\usepackage{bm}                          
\usepackage{epsfig}                      
\usepackage{graphicx}                  
\usepackage{times}
\usepackage{mathtools}
\usepackage{amssymb,amsmath}   

\usepackage{units}
\usepackage{color}

\usepackage{comment}

\usepackage[pagebackref,breaklinks,colorlinks,linkcolor=blue,citecolor=blue,urlcolor=blue,bookmarks=false]{hyperref}
\usepackage{xspace}
\usepackage{setspace}
\usepackage{grfext}
\PrependGraphicsExtensions*{.jpg,.png,.PNG}

\usepackage{gensymb}


%% file: setup/macros.tex
\newcommand{\centeredtab}[1]{\setlength{\tabcolsep}{0pt}\begin{tabular}{c} #1 \end{tabular}}

\usepackage{soul}
\usepackage{svg}

\usepackage[accsupp]{axessibility} 

\DeclareMathAlphabet{\altmathcal}{OMS}{cmsy}{m}{n}
\DeclareMathAlphabet{\mathbfit}{OT1}{ptm}{bx}{it}

\newlength\paramargin
\newlength\figmargin

\newlength\secmargin
\newlength\figcapmargin
\newlength\tabcapmargin

\newcommand {\first}[1]{{\color{red}\textbf{#1}}}
\newcommand {\second}[1]{{\color{blue}\underline{#1}}}

\newenvironment{tightitemize}{
	\vspace{-1.5mm}
	\begin{itemize}
		\setlength{\itemsep}{1pt}
		\setlength{\parskip}{2pt}
		\setlength{\parsep}{0pt}}{\end{itemize}
	
}

\newcommand{\topic}[1]
{
\vspace{1mm}\noindent\textbf{#1}
}

\newcommand{\tabref}[1]{Table~\ref{tab:#1}}

\long\def\ignorethis#1{}

\newbox\jsavebox%

\makeatletter
\newcommand{\providelength}[1]{%
  \@ifundefined{\expandafter\@gobble\string#1}
   {
    \typeout{\string\providelength: making new length \string#1}%
    \newlength{#1}%
   }
   {
    \sdaau@checkforlength{#1}%
   }%
}

\newcommand{\sdaau@checkforlength}[1]{%
  \edef\sdaau@temp{\expandafter\sdaau@getfive\meaning#1TTTTT$}%
  \ifx\sdaau@temp\sdaau@skipstring
    \typeout{\string\providelength: \string#1 already a length}%
  \else
    \@latex@error
      {\string#1 illegal in \string\providelength}
      {\string#1 is defined, but not with \string\newlength}%
  \fi
}
\def\sdaau@getfive#1#2#3#4#5#6${#1#2#3#4#5}
\edef\sdaau@skipstring{\string\skip}
\makeatother

\usepackage[capitalize]{cleveref}
\crefname{section}{Sec.}{Secs.}
\Crefname{section}{Section}{Sections}
\Crefname{table}{Table}{Tables}
\crefname{table}{Tab.}{Tabs.}

\def\c{\mathbf{c}}

\def\t{\mathbf{t}}
\def\x{\mathbf{x}}

\def\xi{\x_{\neg i}}

\def\neg{\hspace{-0.2mm}}

\makeatletter
\newcommand*\MY@rightharpoonupfill@{%
    \arrowfill@\relbar\relbar\rightharpoonup
}
\newcommand*\overrightharpoon{%
    \mathpalette{\overarrow@\MY@rightharpoonupfill@}%
}
\makeatother

\newlength{\depthofsumsign}

%% file: setup/symbols.tex
\def\xi{\mathbf{x}_i}

%% file: setup/graphicspath.tex
\graphicspath{{figures}, {examples}}

%% file: figure/fig_teaser.tex
\begin{center}
\small
\setlength{\fboxrule}{1pt}
\setlength{\tabcolsep}{2pt}
\begin{tabular}{ccc}
\centeredtab{\begin{tikzpicture}
 \foreach \X [count=\Z]in {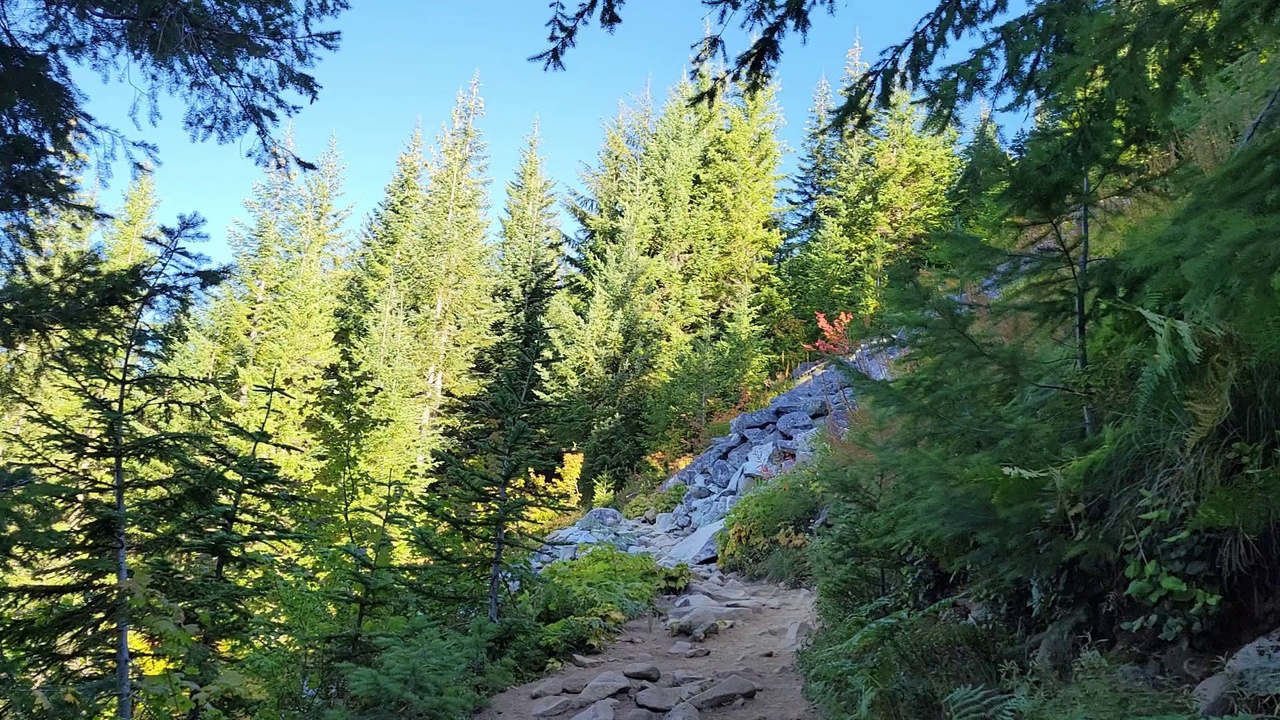,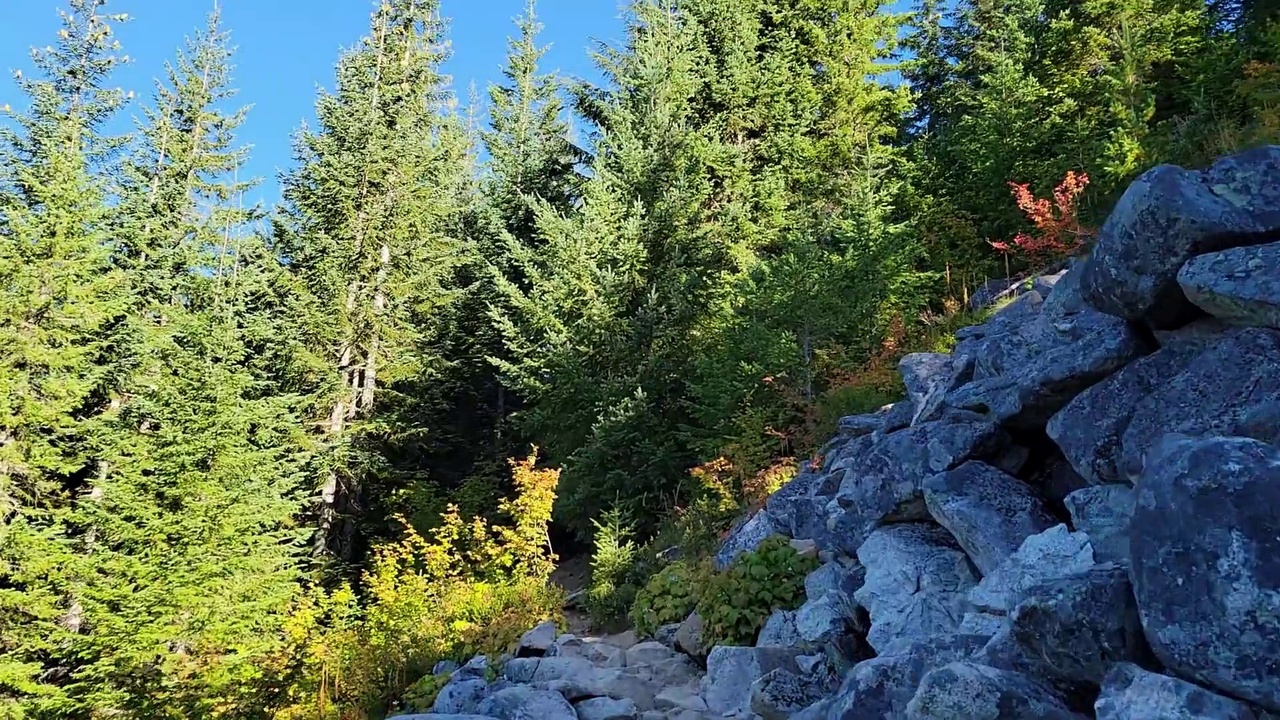,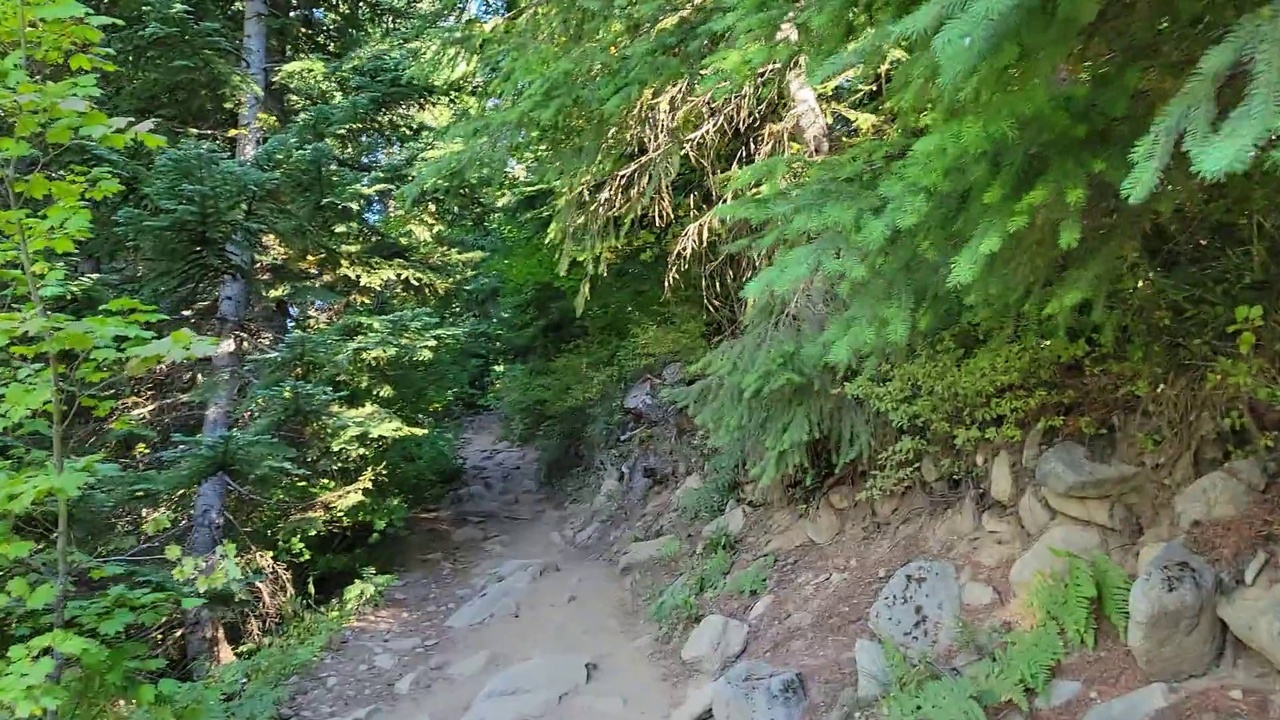,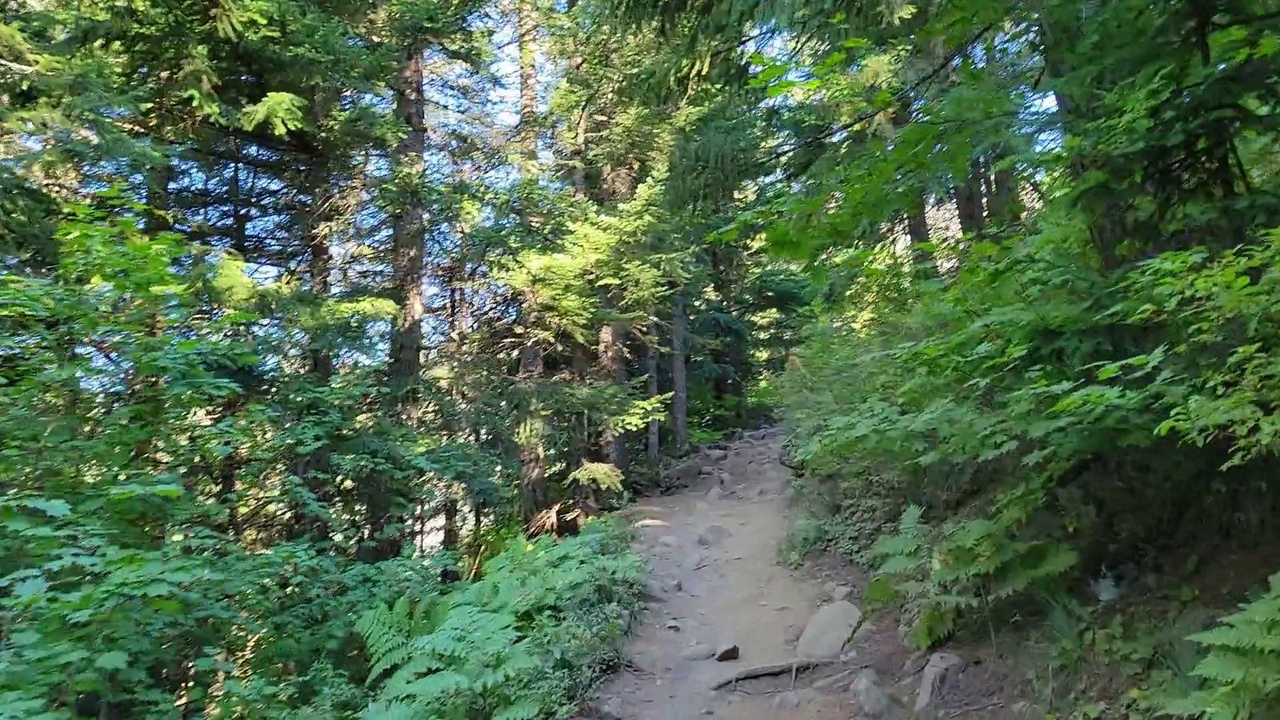,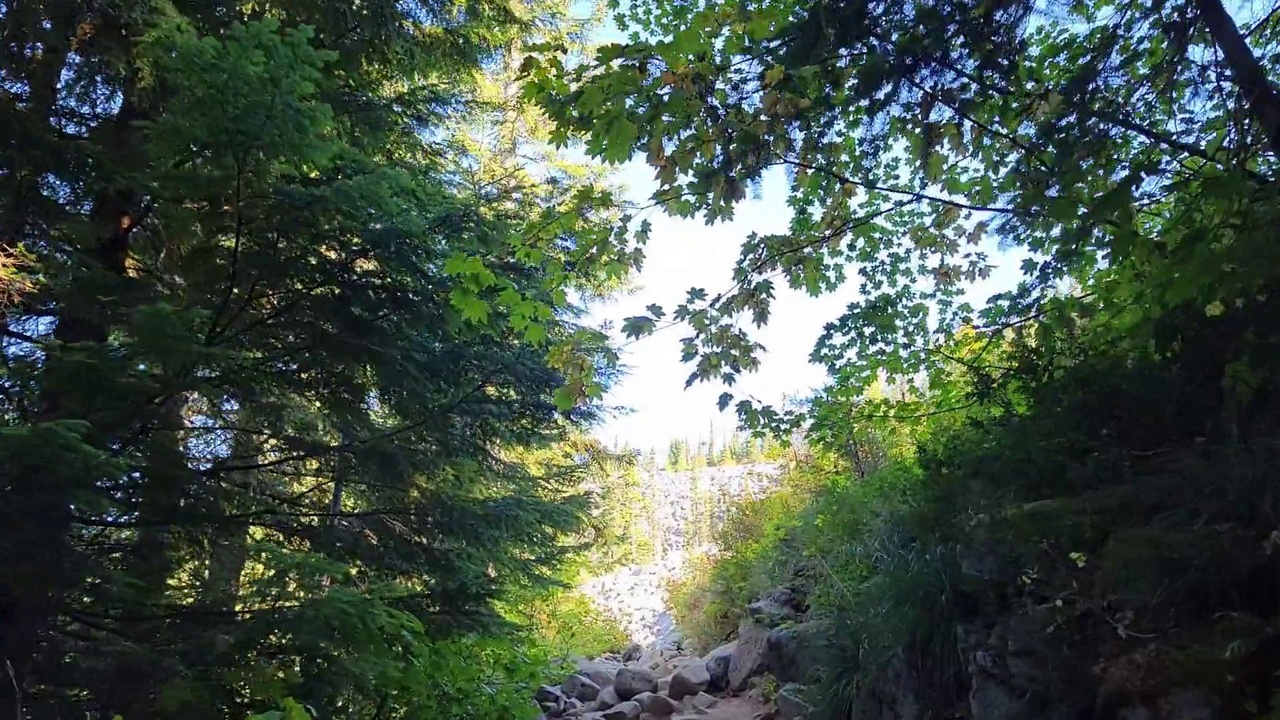,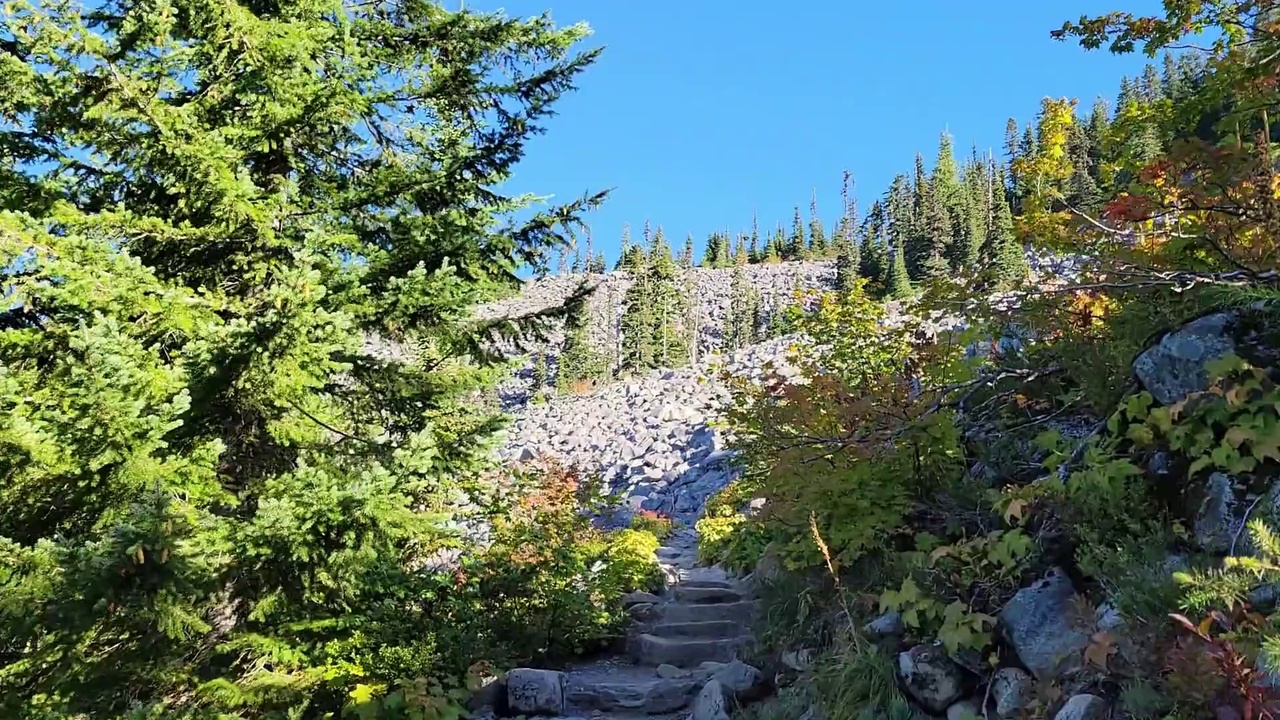,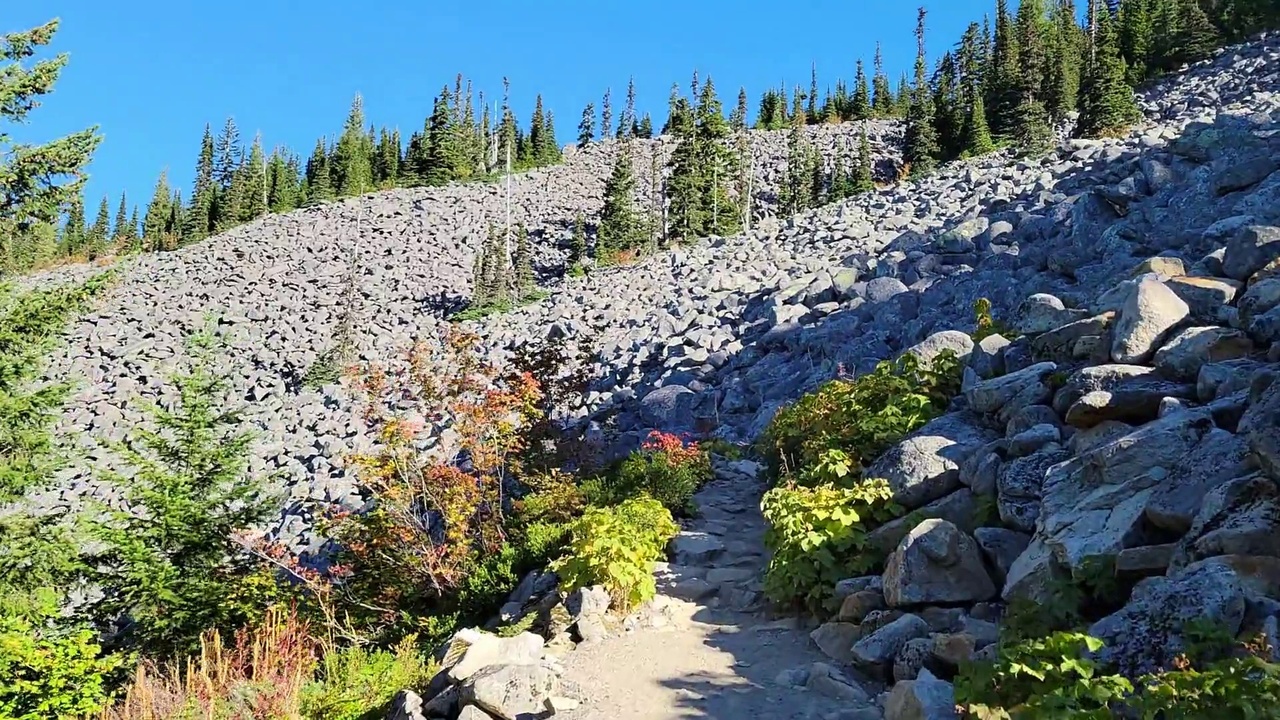}
 {\node[opacity=1] at (\Z/5,-\Z/2.5,0) {\fbox{\includegraphics[width=3cm]{\X}}};}
\end{tikzpicture} \\
Input: casually captured long video \\
\\ \\ \\
\includegraphics[width=5cm]{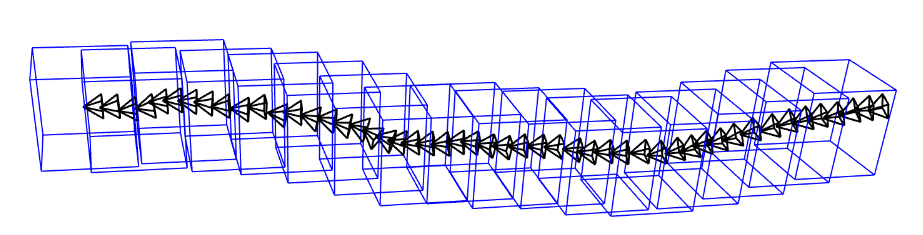} \\
Output: jointly estimated camera poses \\ and local radiance fields
} &
\setlength{\tabcolsep}{1pt}
\renewcommand{\arraystretch}{0.8}
\begin{tabular}{cccc}
{\includegraphics[width=3.9cm]{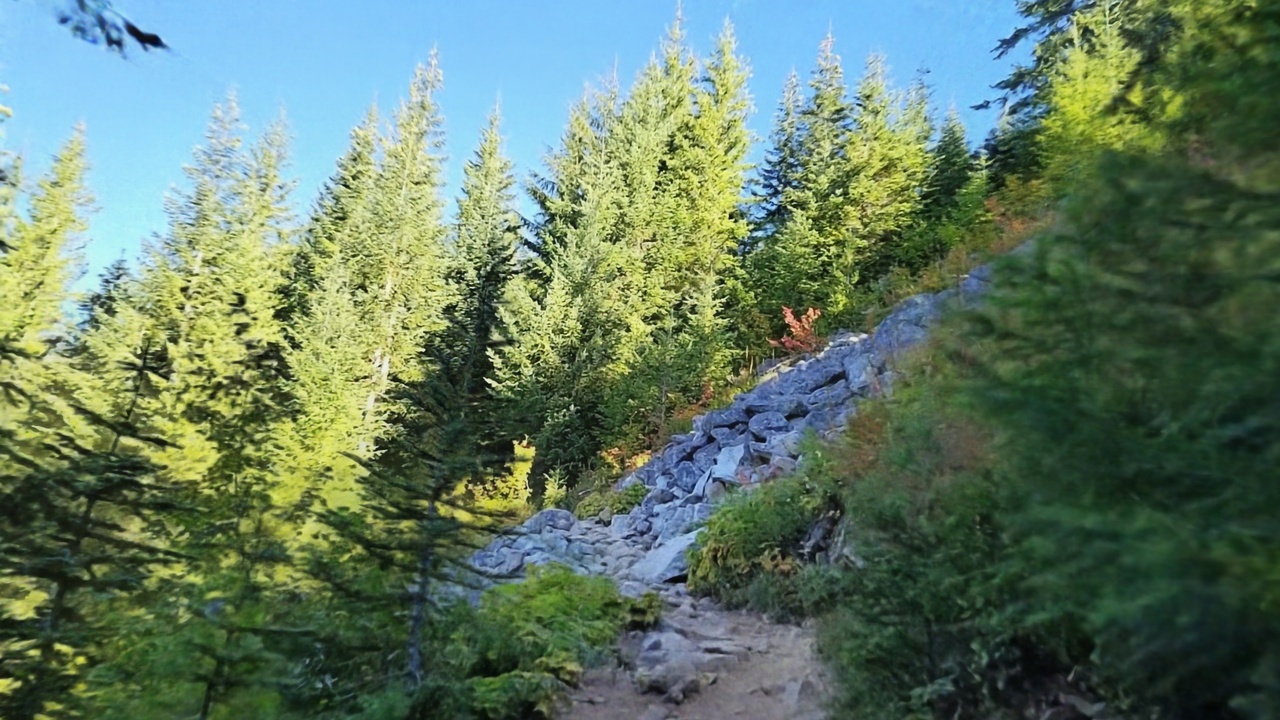}} & 
{\includegraphics[width=3.9cm]{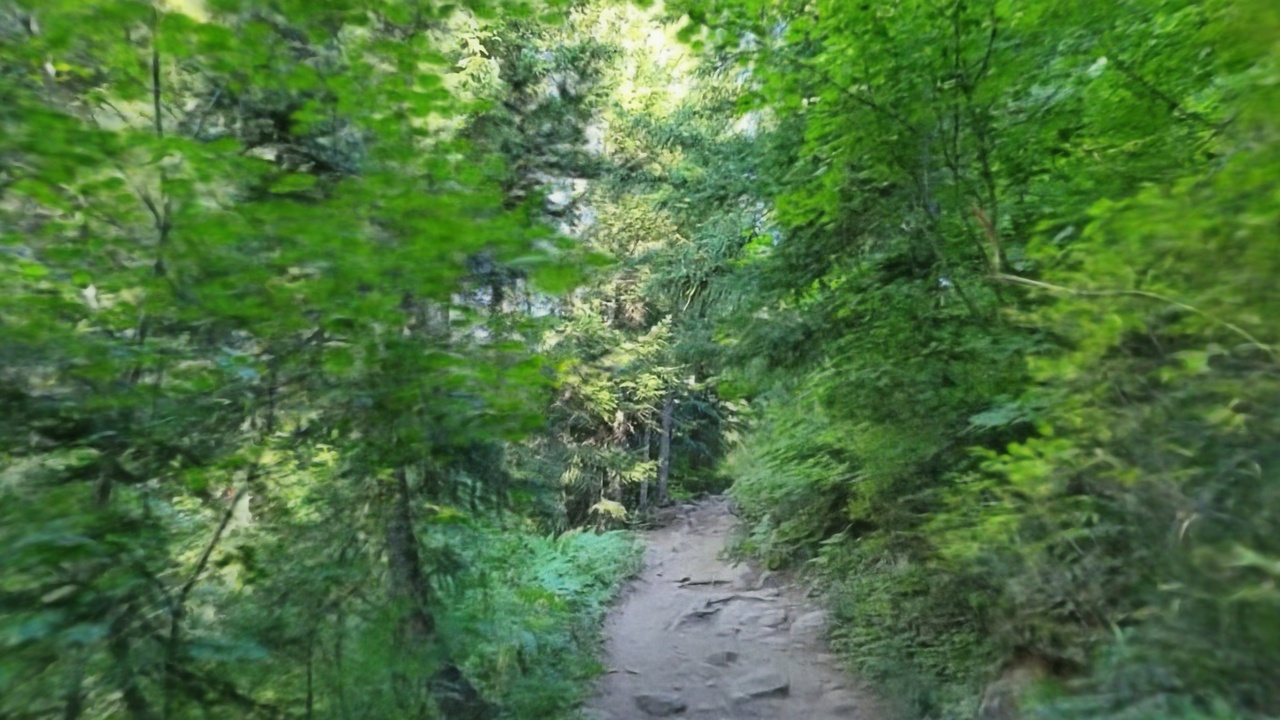}} & 
{\includegraphics[width=3.9cm]{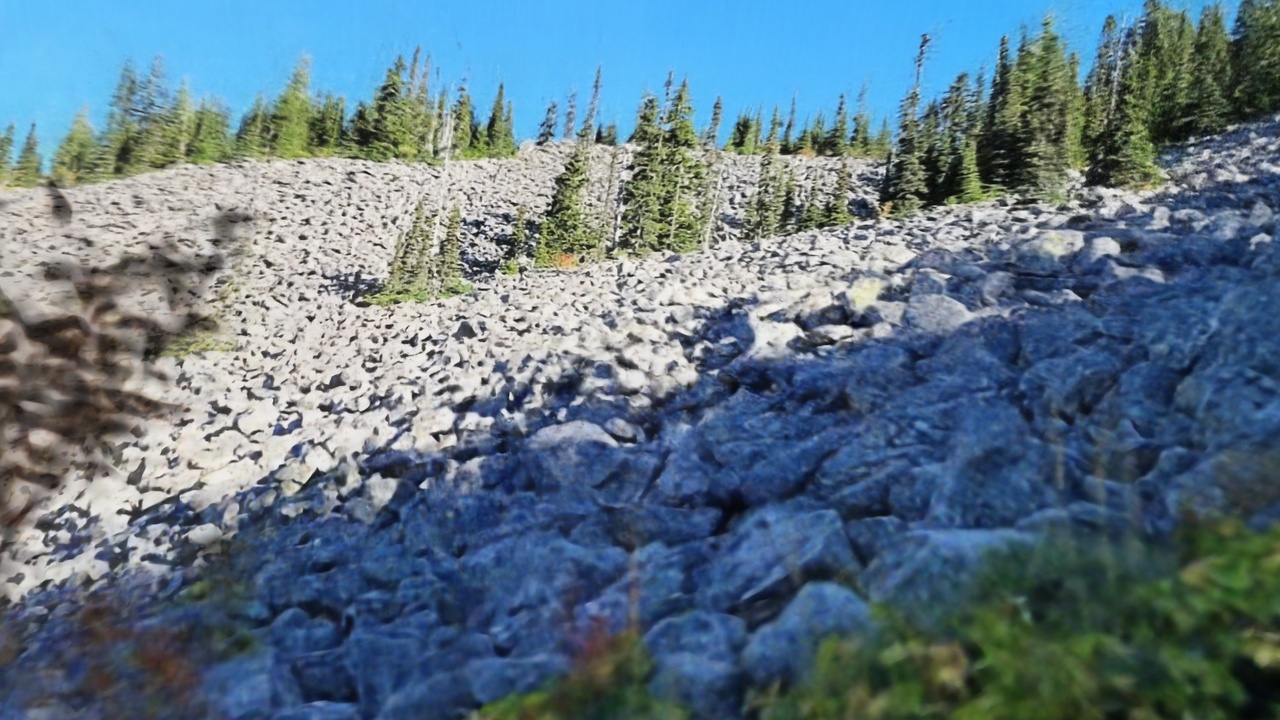}} \\
\multicolumn{3}{c}{LocalRF (ours): high-quality novel view synthesis} \\
\rule{0pt}{2.35cm}{\includegraphics[width=3.9cm]{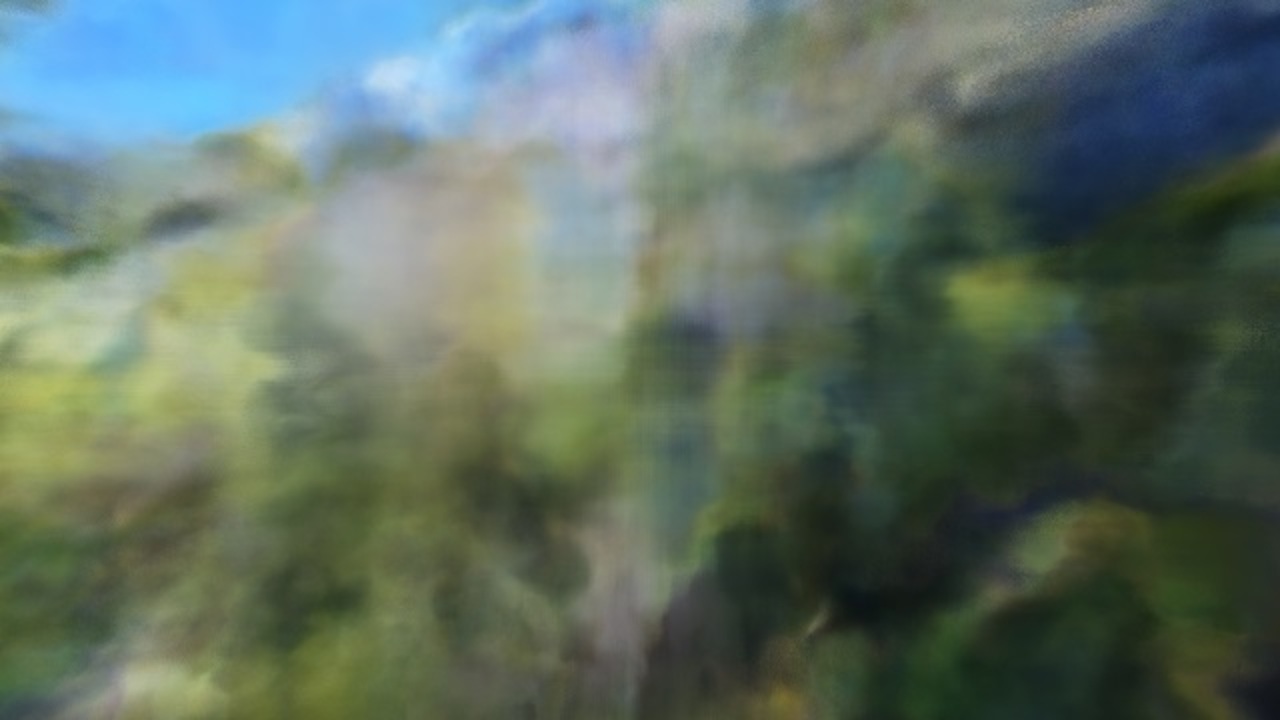}} & 
{\includegraphics[width=3.9cm]{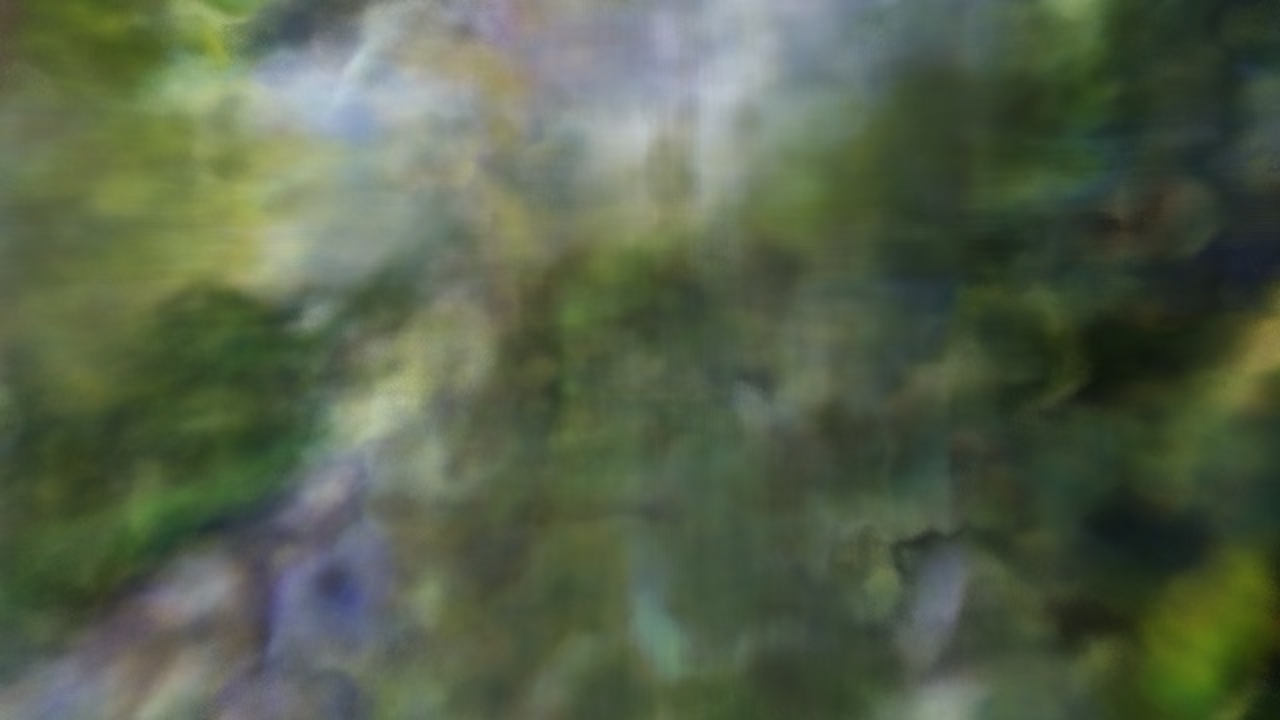}} & 
{\includegraphics[width=3.9cm]{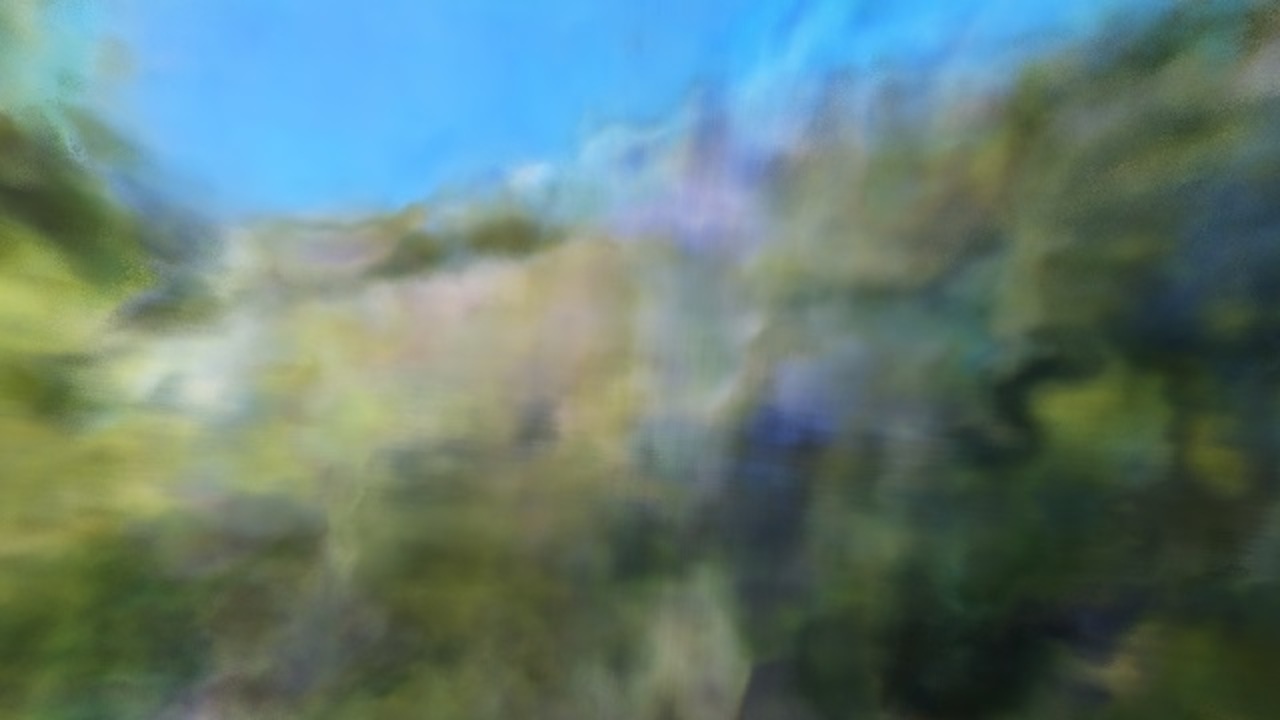}} \\
\multicolumn{3}{c}{BARF~\cite{lin2021barf}: the estimated poses often fall into local minima for long sequences} \\
\rule{0pt}{2.35cm}{\includegraphics[width=3.9cm]{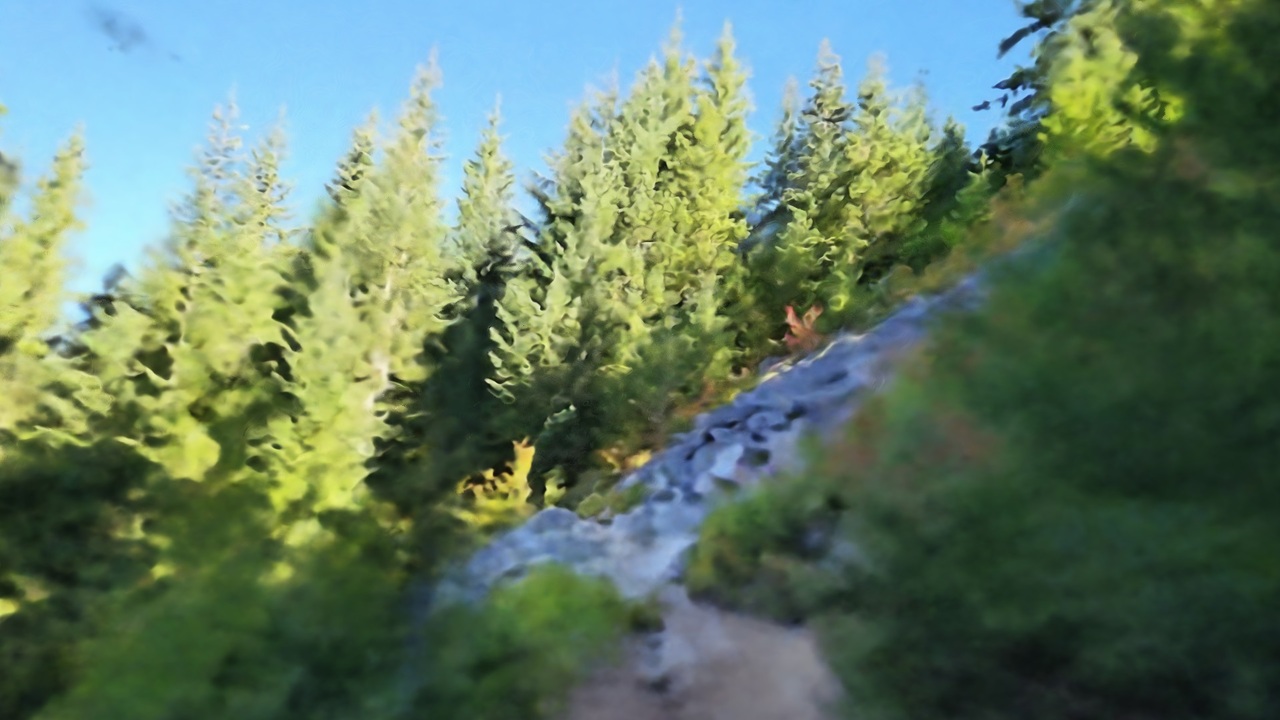}} & 
{\includegraphics[width=3.9cm]{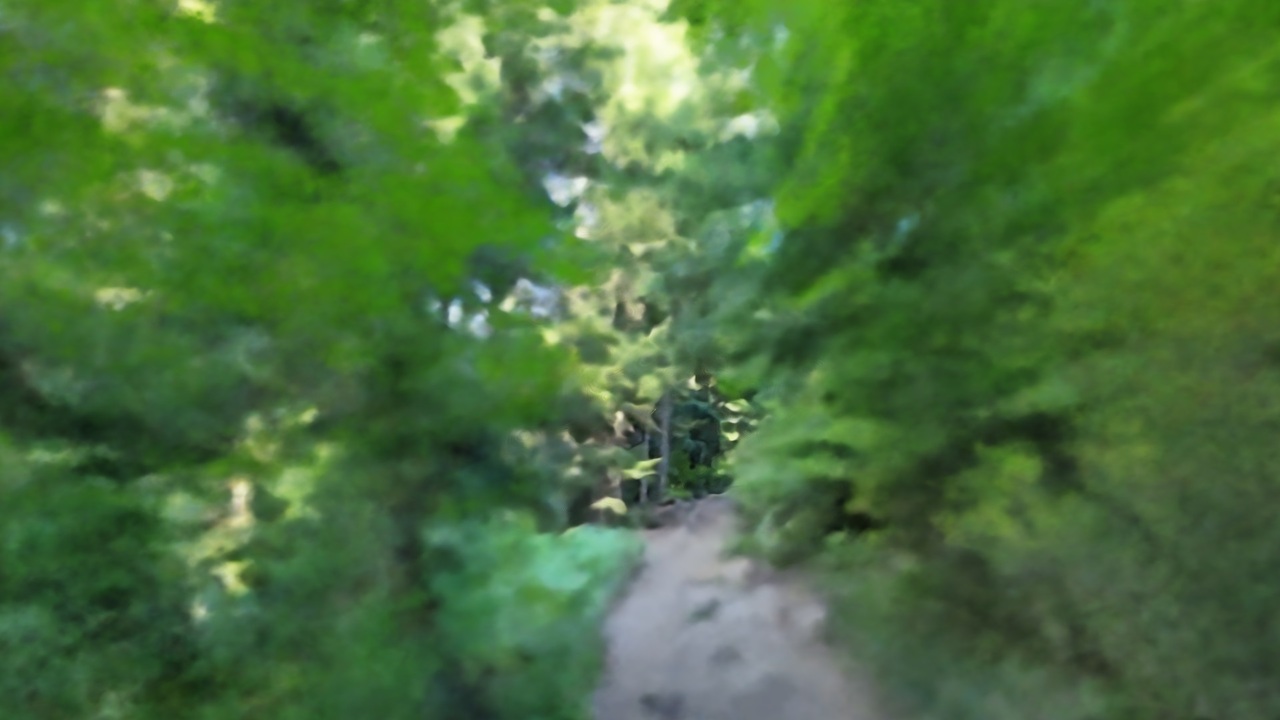}} & 
{\includegraphics[width=3.9cm]{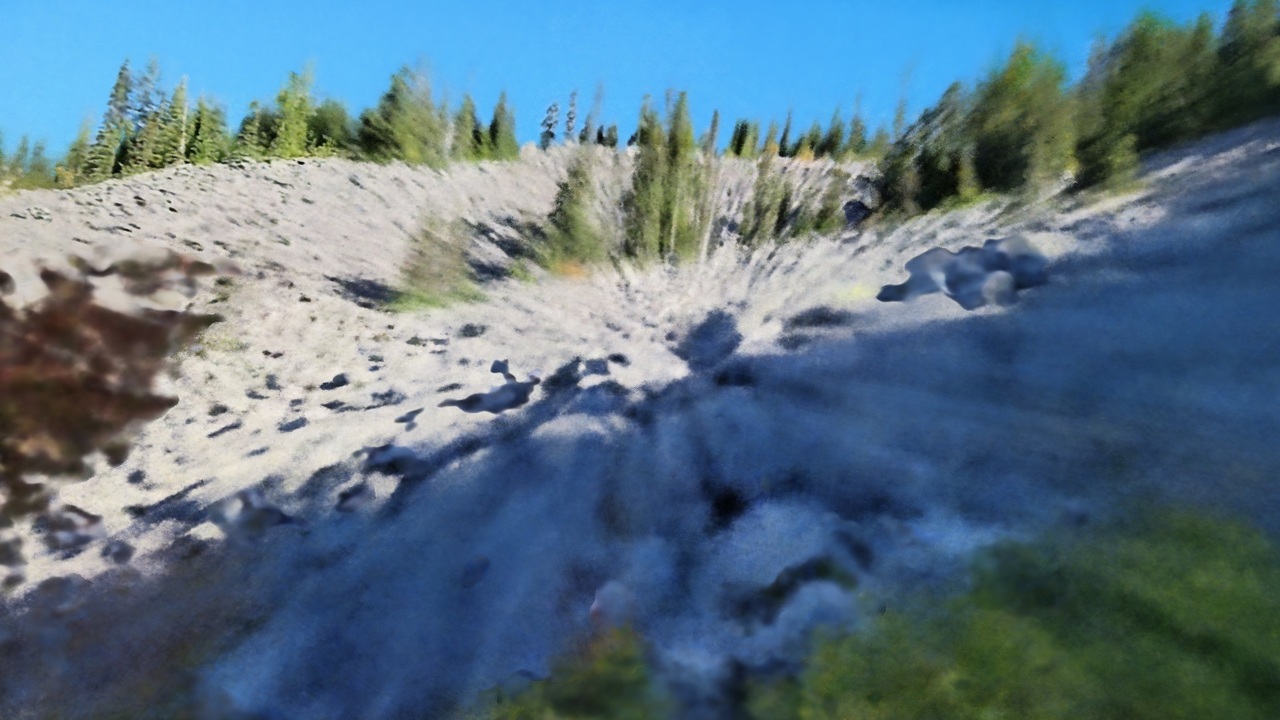}} \\
\multicolumn{3}{c}{Mip-NeRF360~\cite{barron2022mipnerf360}: the spatial resolution is often limited throughout the video}
\end{tabular}
\end{tabular}%
\captionof{figure}{
\label{fig:teaser}
\textbf{High-quality novel view synthesis from a long casually captured video.} 
We jointly optimize camera poses and a scene representation using a progressive scheme that dynamically allocates local radiance fields (blue boxes).
Our method robustly handles casual hand-held captures, scales to processing arbitrarily long videos with limited memory usage, and maintains high resolution throughout the entire video.
}
\end{center}

%% file: 0_abstract.tex
\begin{abstract}
\noindent
We present an algorithm for reconstructing the radiance field of a large-scale scene from a single casually captured video.
The task poses two core challenges.
First, most existing radiance field reconstruction approaches rely on accurate pre-estimated camera poses from Structure-from-Motion algorithms, which frequently fail on in-the-wild videos. 
Second, using a single, global radiance field with finite representational capacity does not scale to longer trajectories in an unbounded scene.
For handling unknown poses, we jointly estimate the camera poses with radiance field in a \emph{progressive} manner.
We show that progressive optimization significantly improves the robustness of the reconstruction.
For handling large unbounded scenes, we dynamically allocate new \emph{local} radiance fields trained with frames within a temporal window.
This further improves robustness (e.g., performs well even under moderate pose drifts) and allows us to scale to large scenes.
Our extensive evaluation on the \textsc{Tanks and Temples} dataset and our collected outdoor dataset, \textsc{Static Hikes}, show that our approach compares favorably with the state-of-the-art.
\end{abstract}

%% file: 1_introduction.tex
\section{Introduction}
\label{sec:intro}
\noindent
Dense scene reconstruction for photorealistic view synthesis has many critical applications, for example, in VR/AR (virtual traveling, preserving of important cultural artifacts), video processing (stabilization and special effects), and mapping (real-estate, human-level maps).
Recently, rapid progress has been made in increasing the fidelity of reconstructions using radiance fields \cite{mildenhall2020nerf}.
Unlike most traditional methods, radiance fields can model common phenomena such as view-dependent appearance, semi-transparency, and intricate micro-details.

\topic{Challenges.} In this paper, we aim to create radiance field reconstructions of \emph{large-scale} scenes that are acquired using a single handheld camera since this is arguably the most practical way of capturing them outside the realm of professional applications.
In this setting, we are faced with two main challenges: 
(1) estimating accurate camera trajectory of a long path and 
(2) reconstructing the large-scale radiance fields of scenes. 
Resolving them together is difficult because changes in observation can be explained by either camera motion or the radiance field's ability to model view-dependent appearance. 
For this reason, many radiance field estimation techniques assume that the accurate poses are known in advance (typically fixed during the radiance field optimization).
However, in practice, this means that one has to use a separate method, such as Structure-from-Motion (SfM), for estimating the camera poses in a pre-processing step.
Unfortunately, SfM is not robust in the handheld \emph{video} setting.
It frequently fails because, unlike radiance fields, it does not model view-dependent appearance and struggles in the absence of highly textured features and in the presence of even slight dynamic motion (such as swaying tree branches).

To remove the dependency on known camera poses, several approaches propose jointly optimizing camera poses and radiance fields~\cite{wang2021nerfmm, lin2021barf, SCNeRF2021}. 
These methods perform well when dealing with a small number of frames and a good pose initialization. 
However, as shown in our experiments, they have difficulty in estimating long trajectories of a video camera from scratch and often fall into local minima.

\topic{Our work.} In this paper, we propose a joint pose and radiance field estimation method.
We design our method by drawing inspiration from classical \emph{incremental SfM} algorithms and \emph{keyframe-based SLAM} systems for improving the robustness.
The core of our approach is to process the video sequence \emph{progressively} using overlapping \emph{local} radiance fields.
More specifically, we progressively estimate the poses of input frames while updating the radiance fields.
To model large-scale unbounded scenes, we dynamically instantiate local radiance fields.
The increased locality and progressive optimization yield several major advantages:
\begin{tightitemize}
\item Our method scales to processing arbitrarily long videos without loss of accuracy and without hitting memory limitations.
\item Increased robustness because the impact of misestimations is locally bounded.
\item Increased sharpness because we use multiple radiance fields to model local details of the scene (see Figure~\ref{fig:teaser} and \ref{fig:mip360vsloc}b).
\end{tightitemize}
We validate our method on the \textsc{Tanks and Temples} dataset.
We also collect a new dataset \textsc{Static Hikes} of twelve outdoor scenes using four consumer cameras to evaluate our method. 
These sequences are challenging due to long handheld camera trajectories, motion blur, and complex appearance. 
We will release the source code for our method and the new dataset for reproducibility.

\topic{Our contributions.} 
We present a new method for reconstructing the radiance field of a large-scale scene, which contains the following contributions:
\begin{tightitemize}
\item We propose to progressively estimate the camera poses and radiance fields, leading to significantly improved robustness. 
\item We show that using multiple overlapping local radiance fields improves visual quality and supports modeling large-scale unbounded scenes.
\item We contribute a newly collected video dataset that presents new challenges not covered by existing view synthesis datasets.
\end{tightitemize}

\topic{Limitations.} 
Our work aims to synthesize novel views from the reconstructed radiance fields. 
While we jointly estimate the poses in the pipeline, we do not perform global bundle adjustment and loop closure (i.e., not a complete SLAM system). 
We leave this important direction for future work.

\input{figure/mip360_vs_local}

%% file: figure/mip360_vs_local.tex
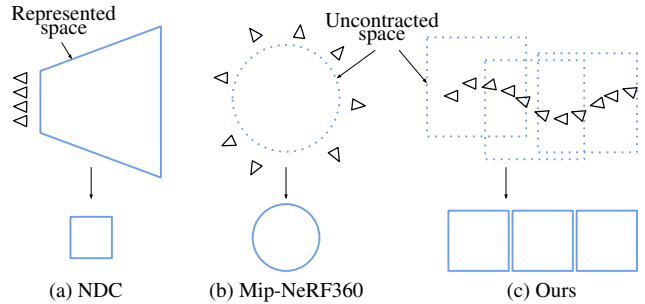
\begin{figure}[t]
\footnotesize
\centering
\begin{overpic}[width=\columnwidth]{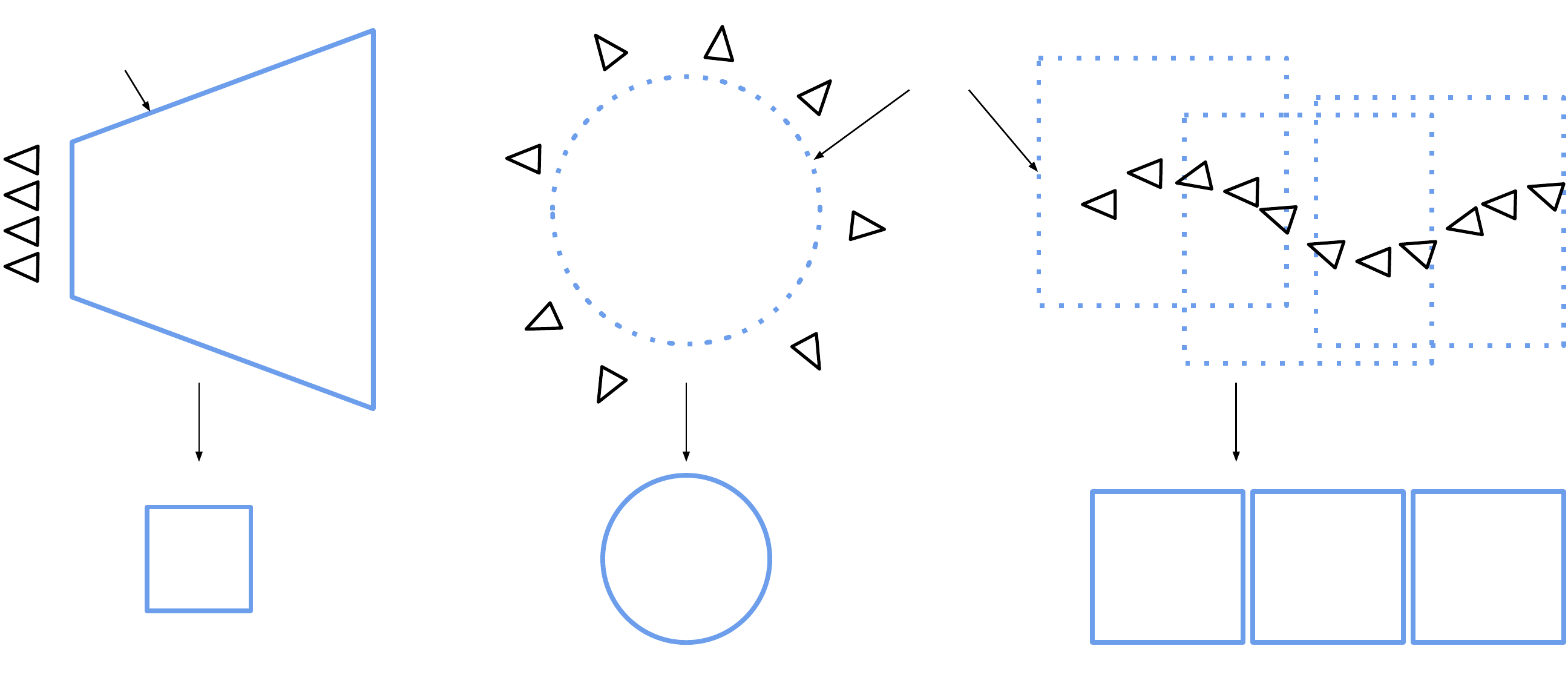}
\put (0, 42) {\renewcommand{\arraystretch}{0}\centeredtab{Represented \\ space}}
\put (50, 41) {\renewcommand{\arraystretch}{0.5}\centeredtab{Uncontracted \\ space}}
\put (6, -1) {(a) NDC}
\put (31.5, -1) {(b) Mip-NeRF360}
\put (79, -1) {(c) Ours}
\end{overpic}
\captionof{figure}{
\textbf{Space parameterization.}
(a) NDC, used by NeRF~\cite{mildenhall2020nerf} for forward-facing scenes, maps a frustum to a unit cube volume. 
While a sensible approach for forward-facing cameras, it is only able to represent a small portion of a scene as the frustum cannot be extended beyond a field of view of $120^\circ$ or so without significant distortion. 
(b) Mip-NeRF360's~\cite{barron2022mipnerf360} space contraction squeezes the background and fits the entire space into a sphere of radius 2. 
It is designed for inward-facing 360 scenes and cannot scale to long trajectories. 
(c) Our approach allocates several radiance fields along the camera trajectory. 
Each radiance field maps the entire space to a $[-2, 2]$ cube (Equation~\eqref{eq:contract}) and, each having its own center for contraction (Equatione~\eqref{eq:shift}), the high-resolution uncontracted space follows the camera trajectory and our approach can adapt to any camera path. 
}
\label{fig:mip360vsloc}
\vspace{-3mm}
\end{figure}

%% file: 2_related.tex
\section{Related Work}
\label{sec:related}

\topic{Novel view synthesis.} 
Novel view synthesis aims to synthesize new views from multiple posed images.
Recently, neural implicit representation has shown promising novel view synthesis results \cite{mildenhall2020nerf}. 
However, achieving high-quality artifact-free rendering results is still a challenging task.
Recent works further improve the visual quality by addressing inconsistent camera exposure or illumination~\cite{rematas2022urf,tancik2022blocknerf,martinbrualla2020nerfw}, handling dynamic elements~\cite{xian2021space,li2021neural,Gao2021DynNeRF,park2021nerfies,park2021hypernerf,liu2023robust}, anti-aliasing~\cite{barron2021mipnerf}, high noise~\cite{mildenhall2021rawnerf} or optimization from a reduced number of frames~\cite{Niemeyer2021Regnerf}.
While these implicit representation-based methods yield high-quality results, they take days to train.
To improve the training efficiency, some works also explore more explicit representations with voxel-like structures \cite{yu_and_fridovichkeil2021plenoxels,Sun_2022_CVPR}, tensor factorization \cite{tensorf}, light field representation~\cite{attal2022learning,attal2023hyperreel}, or hashed voxel/MLP hybrid~\cite{mueller2022instant}. 
Our work also leverages the recent advantage of TensoRF~\cite{tensorf}.

\topic{Scalable view synthesis.} 
Several systems have been proposed to support unbounded scenes \cite{barron2022mipnerf360,kaizhang2020nerfpp}.
However, these methods require either omnidirectional inputs \cite{jang2022egocentric}, proxy geometry \cite{wu2022snisr}, specialized drone shots \cite{Turki_2022_meganerf}, or satellite shots \cite{xiangli2022bungeenerf} and struggle with monocular videos captured at ground level.
Recently, Mip-NeRF 360~\cite{barron2022mipnerf360} contracts background into a contracted space, and NeRF++~\cite{kaizhang2020nerfpp} optimizes an environment map to represent the background. 
BlockNeRF~\cite{tancik2022blocknerf} is scalable but requires multiview inputs and several observations.
NeRFusion~\cite{zhang2022nerfusion} constructs per frame local feature volumes using a pretrained 2D CNN followed by a sparse 3D CNN. 
It is scalable and has demonstrated good accuracy on large indoor scenes, but it does not tackle camera pose estimation or unbounded outdoor scenes. 
Since the representation is mostly reconstructed before an optional per-scene optimization, it is not trivial to optimize poses simultaneously.

In contrast to all these constraints, our method is robust, works with arbitrary long camera trajectories, and only takes casually captured monocular first-person videos as input.

\topic{Camera pose estimation.}
Visual odometry estimates camera poses from videos. They can either rely directly on the color by maximizing the photoconsistency~\cite{zhou2017unsupervised,yin2018geonet} or on extracted hand-crafted features~\cite{mur2015orb,mur2017orb,schoenberger2016sfm}.
Recently, learning-based methods~\cite{zhou2017unsupervised,godard2019digging,kopf2021robust,teed2021droid,zhao2022particlesfm} learn to optimize the camera trajectories in a self-supervised way and show strong results.
Similarly, many methods extend NeRF to optimize the camera poses jointly with radiance fields from photometric loss~\cite{wang2021nerfmm, lin2021barf, SCNeRF2021}.
However, these methods struggle to reconstruct and synthesize faithful images for large scenes and often fail for monocular first-person videos with long camera trajectories.
Vox-Fusion~\cite{yang2022vox} and Nice-SLAM~\cite{Zhu2022CVPR} achieve good pose estimation but are designed for RGB-D inputs and require accurate depth: Vox-Fusion to allocate a sparse voxel grid and Nice-SLAM to determine where to sample along the ray.
Note that our goal does \emph{not} lie in estimating camera poses. 
Instead, focus on reconstructing overlapping local radiance fields that enable photorealistic view synthesis. 
We believe integrating advanced techniques such as global bundle adjustment can improve our results.

%% file: 3_method.tex
\input{figure/fig_overview}
\section{Method}
\label{sec:method}
\noindent
Our method takes a potentially very long monocular video of a large-scale scene as input.
Our goal is to reconstruct the radiance field of the scene along with the camera trajectory to enable free-viewpoint novel view synthesis. 

We choose TensoRF~\cite{tensorf} as our base representation for its quality, reasonable training speed and model size. 
TensoRF models the scene with a factorized 4D tensor that maps a 3D position $\x$ to the corresponding volume density $\sigma$ and view-dependent color $\c$.
High-quality novel view synthesis results for small-scale scenes have been demonstrated using this representation.
However, it has only been achieved with accurate pre-known camera poses, and TensoRF's representation power needs to be increased for capturing the details from long trajectories of unbounded scenes.

In this work, we resolve the need for pre-known camera poses by improving the \emph{robustness} of joint camera pose and radiance field estimation, and we \emph{scale the method} to handle arbitrarily long input sequences. 
To this end, we propose a progressive optimization scheme that processes the input video with a sweeping temporal window and incrementally updates the radiance fields and the camera poses.
This process ensures that new frames are added to a well-converged solution for camera poses and radiance field representation of the previous structures, effectively preventing getting stuck in poor local minima.
In addition, we dynamically allocate new \emph{local} radiance fields throughout the optimization that are supervised by a limited number of input frames (within a temporal window). 
This further improves robustness while processing arbitrarily long videos with fixed memory. 

\subsection{Formulation and Preliminaries}
\noindent
During our optimization procedure, we estimate $P$ camera poses $[R|t]_k, k \in [1..P]$\footnote{We use the continuous 6D representation~\cite{Zhou_2019_6D} for representing camera rotations.} as well as the parameters of a series of $M$ local radiance fields $\Theta_j, j \in [1..M]$. 

Given a pixel, we use the camera parameters and the poses to generate a ray $\mathbf{r}$. Along this ray, we sample 3D positions $\{\x_i\}$ and query a radiance field that provides color and density:
\begin{equation}
(\mathbf{c}_i, \sigma_i) = \text{RF}_{\Theta_j}(\x_i).
\end{equation}
Via volume rendering~\cite{kajiya1984ray,drebin1988volume}, we can render the ray using this representation:
\begin{align}
\label{eq:volume_rendering}
\hat{\mathbf{C}}(\mathbf{r}) &= \sum_{i=1}^{N} T_i(1-\text{exp}(-\sigma_i\delta_i))\mathbf{c}_i,\\
T_i &= \text{exp}\left(-\sum_{i-1}^{j}\sigma_j\delta_j\right),
\end{align}
where $\delta_i$ is the distance between two consecutive sample points and $N$ is the number of samples along the ray, and $T_i$ indicates the accumulated transmittance along the ray. 
We optimize the radiance field parameters $\Theta_j$ and the camera pose used to generate the ray using the input frame's color $\mathbf{C}$ as supervision:
\begin{equation}
\mathcal{L} = \left \| \hat{\mathbf{C}}(\mathbf{r}) - \mathbf{C}(\mathbf{r}) \right \|_{2}^{2}.
\label{eq:photometric_loss}
\end{equation}
Using TensoRF~\cite{tensorf} in this context is particularly suitable as it features an explicit coarse-to-fine optimization analogous to BARF's~\cite{lin2021barf} and reduces the likelihood of converging to a local minimum for pose estimation.

To handle unbounded scenes, we leverage a scene parameterization similar to Mip-NeRF360~\cite{barron2022mipnerf360}'s contraction, mapping every point to a $[-2, 2]$ space before querying our radiance field model:
\begin{equation}
    \label{eq:contract}
    \text{contract}(\x) = 
    \begin{cases}
      \x & \text{if}\ \| \x \|_{\infty} \leq 1 \\
      \left(2 - \frac{1}{\| \x \|_{\infty}}\right) \left(\frac{\x}{\| \x \|_{\infty}}\right) & \text{otherwise}.
    \end{cases}
\end{equation}
Here we use the $L_{\infty}$ norm to fully utilize TensoRF's square bounding boxes.
While Mip-NeRF360 scale camera poses to keep the uncontracted space around the area of interest, we cannot adopt this strategy since we jointly estimate poses and radiance fields (the poses are unknown a priori).
We achieve appropriate scaling by dynamically creating new radiance fields (see Figure~\ref{fig:mip360vsloc} and Section~\ref{sec:local}).

\subsection{Progressive Joint Camera Pose and Radiance Field Optimization}
\label{sec:prog}
\input{figure/fig_prog_impact}
\noindent
Existing pose-calibrating methods~\cite{lin2021barf,wang2021nerfmm,SCNeRF2021} have demonstrated that jointly optimizing a radiance field and camera poses can achieve satisfactory results in small-scale scenes. 
However, when dealing with longer sequences, joint optimization fails as estimated poses get stuck in local minima (see Figure~\ref{fig:prog_impact}). 
To improve the robustness, we start the optimization process with only a small number of frames (the first five frames in our experiments), and from there we \emph{progressively} introduce subsequent frames to the optimization. 
To this end, we initialize the new pose (index $p + 1$) using the current frame at the end of the trajectory:
\begin{equation}
    [R|t]_{p + 1} \leftarrow [R|t]_p.
\end{equation}
We then add $[R|t]_{p + 1}$ to the trainable parameters and we add the frame's color as supervision of the radiance fields.
In this scheme, the convergence of the parameters for the new frame benefits from the initialization of the radiance field and the currently estimated poses, making it less prone to get stuck in local minima. 
Since we add the camera pose at the end of the trajectory, it also introduces a locality prior that enforces each pose to be close to the former one without an explicit constraint, as it is common for videos.

To further constrain this complex joint optimization problem, we introduce additional losses described in Section~\ref{sec:implementation}.

\subsection{Local Radiance Fields}
\label{sec:local}
\input{figure/fig_loc_impact}
\noindent
The progressive scheme proposed in the previous section provides more robust pose estimation, but it still relies on a single global representation of the scene, which causes problems when modeling long videos:
(1) any misestimation (e.g., outlier pose) has global impact and might cause the entire reconstruction to break down.
(2) a single model with fixed capacity cannot represent arbitrarily long videos with an arbitrary amount of detail, leading to blurry renderings (Figure~\ref{fig:loc_impact}b).
A natural solution to these problems would be to pre-partition the space using radiance field tiling similar to Mega-NeRF~\cite{Turki_2022_meganerf}. 
However, this approach is not applicable in our setting because the camera poses are unknown before the optimization.
To resolve this issue, we dynamically create a new radiance field, whenever the estimated camera pose trajectory leaves the uncontracted space of the current radiance field.
We centered the new radiance field at the location $\t_j$ of the last estimated camera pose: $\t_j \leftarrow t_p$ (Figure~\ref{fig:overview}d.
When sampling a ray, we use this translation to center the radiance fields:
\begin{equation}
    \label{eq:shift}
    (\mathbf{c}_i, \sigma_i) = \text{RF}_{\Theta_j}(\x_i - \t_j).
\end{equation}
We supervise each radiance field with a local subset of the video frames.
This subset contains all the frames during which the radiance field was current, as well as the preceding 30 frames to provide for some overlap.
The overlap is important for achieving consistent reconstructions in the local radiance fields.
We further increase consistency by blending together the rendered colors $\hat{\mathbf{C}_j}(\mathbf{r})$ of all overlapping radiance fields at any supervising frame.
We use per-frame blending weights that increase/decrease linearly within the overlap region.
When we create a new radiance field we stop optimizing previous ones (i.e., freeze them).
At this point we can clear any supervising frames from memory that are not needed anymore.

With this procedure, each radiance field is supervised using only a subset of the frames, granting more robustness. 
The high-resolution space follows the camera trajectory, allowing for more sharpness and scalability.
Figure~\ref{fig:mip360vsloc} shows that local radiance fields help maintain the high-quality uncontracted space around the camera trajectory, which allows for a sharper representation.

\subsection{Implementation}
\label{sec:implementation}
\vspace{-2mm}
\topic{Losses.}
In addition to the supervision from the input color in Equation~\eqref{eq:photometric_loss}, we add monocular depth and optical flow between neighbor frames as they have proven their capabilities to improve the optimization stability in challenging scenarios~\cite{Luo2020cvd,Gao2021DynNeRF}. 
We use RAFT~\cite{teed2020raft} to estimate the frame-to-frame optical flow $\mathcal{F}_{k\rightarrow k+1}, k\in[1..P-1]$ and DPT~\cite{Ranftl2021dpt2} to estimate the per-frame monocular depth $\mathbf{D}$. 
To enable these losses, we first render the depth maps by swapping the sample's color $\c_i$ by the sample's distance to the ray origin $d_i$ in Equation~\eqref{eq:volume_rendering}:
\begin{equation}
    \label{eq:depth_render}
    \hat{\mathbf{D}}(\mathbf{r}) = \sum_{i=1}^{N} T_i(1-\text{exp}(-\sigma_i\delta_i))d_i.
\end{equation}
We formulate depth supervision following the shift and scale invariant loss typically used for monocular depth training~\cite{Ranftl2020dpt1, Ranftl2021dpt2} and radiance fields supervision~\cite{Gao2021DynNeRF}:
\begin{equation}
    \label{eq:depth_loss}
    \mathcal{L}_d = \left| {\hat{\mathbf{D}}^*} - {\mathbf{D}^*}\right|,
\end{equation}
where $\hat{\mathbf{D}}^*$ and $\mathbf{D}^*$ are per-frame normalized depth since monocular depth is not scale and shift invariant. We normalize following~\cite{Ranftl2020dpt1}, by first estimating scale and shift:
\begin{equation}
    \label{eq:depth_normalization0}
    t(\mathbf{D}) = \text{median}(\mathbf{D}),\ s(\mathbf{D}) = \frac{1}{M} \sum_1^M |\mathbf{D} - t(\mathbf{D})|,
\end{equation}
where we have $M$ samples for the frame. Then, we normalize:
\begin{equation}
    \label{eq:depth_normalization}
    \mathbf{D}^* = \frac{\mathbf{D} - t(\mathbf{D})}{s(\mathbf{D})}.
\end{equation}
In practice, we sample rays from 16 images in each batch and obtain scale and shift for each.
To obtain the expected optical flow from our representation, we leverage the relative camera poses and the rendered depth map:
\begin{equation}
    \label{eq:flow_render}
    \hat{\mathcal{F}}_{k\rightarrow k+1} = (u,v) - \Pi\left([R|t]_{k\rightarrow k+1}\Pi^{-1}(u,v,\hat{D})\right)
\end{equation}
Where $\Pi$ projects a 3D point to image coordinates and $\Pi^{-1}$ unprojects a pixel coordinate and depth into a 3D point and $[R|t]_{k\rightarrow k+1} = [R|t]_k^{-1}[R|t]_{k+1}$ is the relative camera pose between the two consecutive frames, bringing a point in the $k^{\text{th}}$ camera's space to the $k+1^{\text{th}}$'s. 
We finally compare the predicted flow to the expected flow from the representation:
\begin{equation}
    \label{eq:flow_loss}
    \mathcal{L}_f = \left\| \hat{\mathcal{F}}_{k\rightarrow k+1} + \mathcal{F}_{k\rightarrow k+1}  \right\|_1
\end{equation}
We use the same process to supervise using the backward optical flow $\mathcal{F}_{k\rightarrow k-1}$.
Note that optical flow computations leverage directly poses and the geometry of the scene in Equation~\eqref{eq:flow_render}, which gives a clear gradient signal for their optimization.

\topic{Scheduling.}
All parameters are optimized using Adam~\cite{2015adam} with $\beta_1=0.9$ and $\beta_1=0.99$.
We start with five poses initialized as identity and an initial TensoRF model. 
Then, for every 100 iterations, we add the next supervising frame in the video as described in Section~\ref{sec:prog} and Figure~\ref{fig:overview}a and \ref{fig:overview}b. 
During this optimization process, we maintain all learning rates, loss weights, and TensoRF resolution at their initial states. 
This ensures that the radiance field does \emph{not} overfit to the first frames. 
Our initial learning rates are $5\cdot10^{-3}$ for rotations and $5\cdot10^{-4}$ for translations, initial TensoRF resolution is $64^3$ and initial regularizing loss weights are $1$ for flow loss and $0.1$ for depth. 
We proceed with the progressive frame registration until an estimated camera pose is beyond the uncontracted space: $\| t_{p} - \t_{j} \|_{\infty} \geq 1$, where $t_{p}$ the last registered frame's translation and $\t_{j}$ the currently optimizing radiance fields center. 
From this point, we refine the TensoRF, and the camera poses for $600$ iterations per add frame (Figure~\ref{fig:overview}c). 
The schedulers and the regularizing losses follow an exponential decrease towards a $0.1$ factor, and we upsample the TensoRF up to $640^3$.
After this stage, we allocate a new TensoRF following Section~\ref{sec:local} and Figure~\ref{fig:overview}d and disable the supervision from the first frames. 
We repeat this process until the entire trajectory is reconstructed.
The optimization takes 30 to 40 hours for 1000 frames on a single NVIDIA TITAN RTX.

%% file: figure/fig_overview.tex
\begin{figure*}[t]
\renewcommand{\arraystretch}{-0.5}
\begin{tabular}{l}
\begin{overpic}[width=1\linewidth]{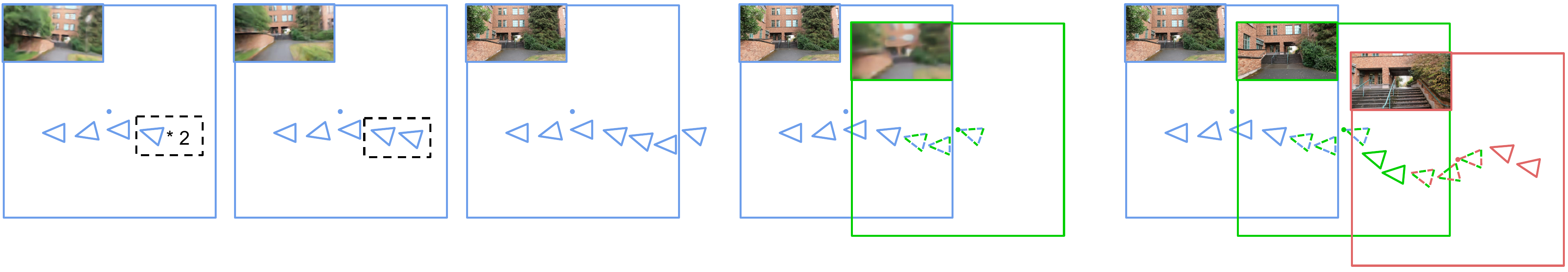}
\put (1, 1)  {\footnotesize (a) Append frame}
\put (16, 0.8) {\footnotesize \centeredtab{(b) Optimize RF \\ and poses}}
\put (32.3, 0.8) {\footnotesize \centeredtab{(c) Refine RF \\ and poses}}
\put (46.5, 0.3) {\footnotesize (d) Allocate another radiance field}
\put (75, -1.5) {\footnotesize (e) Complete reconstruction}
\end{overpic} \\
$\underbrace{\hspace{0.29\linewidth}}_{\text{\footnotesize Repeat until out of uncontracted bound}}$  \\
$\underbrace{\hspace{0.70\linewidth}}_{\text{\footnotesize Repeat until all frames are registered}}$
\end{tabular}
\vspace{\figcapmargin}
\vspace{-1mm}
\caption{\textbf{Method overview.} 
The squares represent the uncontracted space of each local radiance field and the triangles are camera poses.
The color of each camera pose indicates to which radiance fields it is linked and serves as supervision.
We show as insert the renders intermediate at the intermediate optimization step for each local radiance field. (a) We add a frame at the end of the trajectory before (b) jointly estimating poses and the corresponding local radiance field. After the pose reaches the boundary of the high-resolution uncontracted space, (c) we run the optimization without adding frames to refine the poses and the radiance field. Then, to dynamically extend the representation, (d) we allocate a new radiance field. We repeat this process until we cover the full trajectory to produce (e) a complete reconstruction.  
}
\label{fig:overview}
\vspace{-3mm}
\end{figure*}

%% file: figure/fig_prog_impact.tex
\begin{figure}[t]
\centering
\small
\renewcommand{\arraystretch}{0.6}
\begin{tabular}{cc}
\includegraphics[width=0.46\columnwidth]{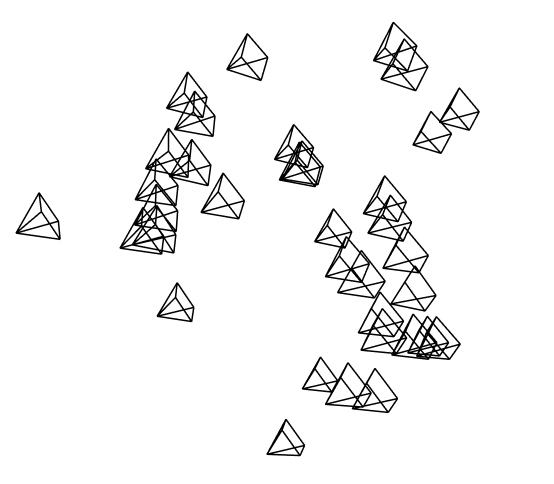} & \includegraphics[width=0.46\columnwidth]{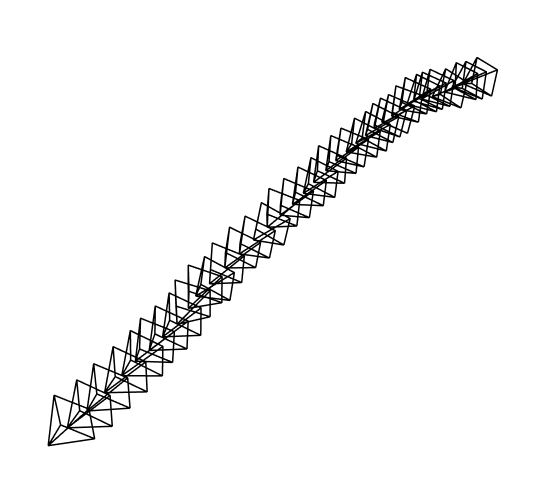} \\
\includegraphics[width=0.46\columnwidth]{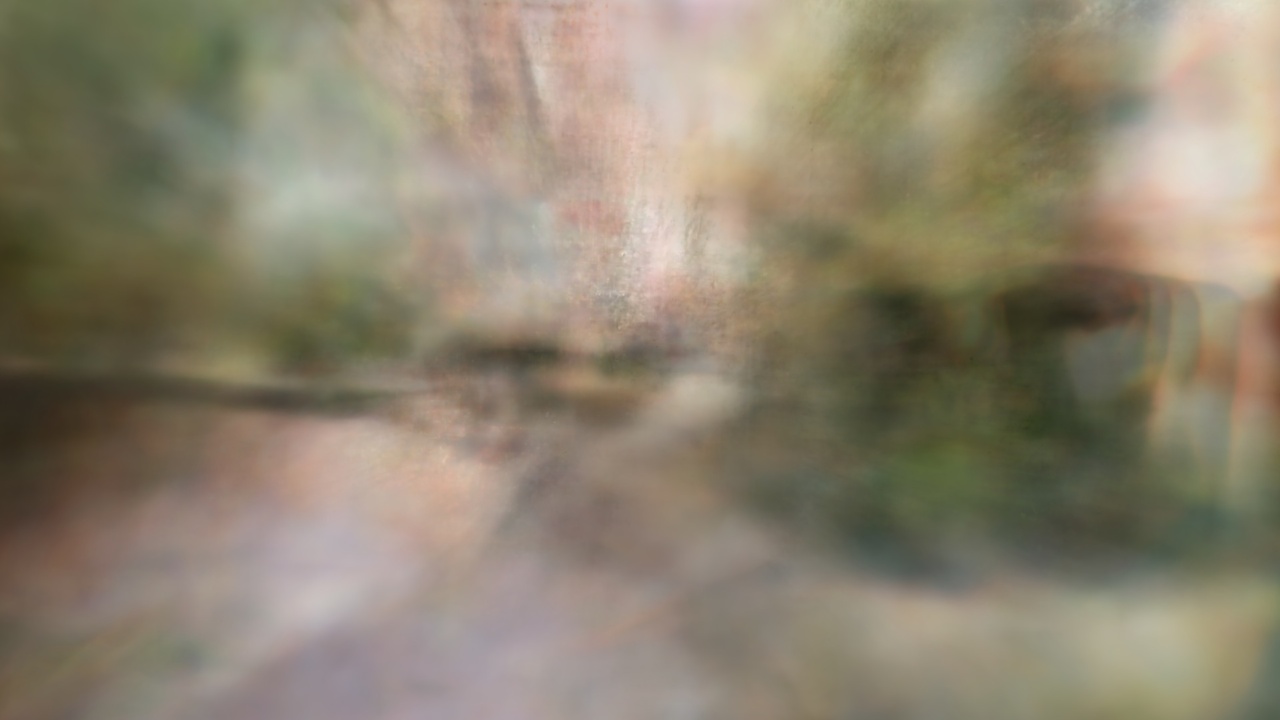} & \includegraphics[width=0.46\columnwidth]{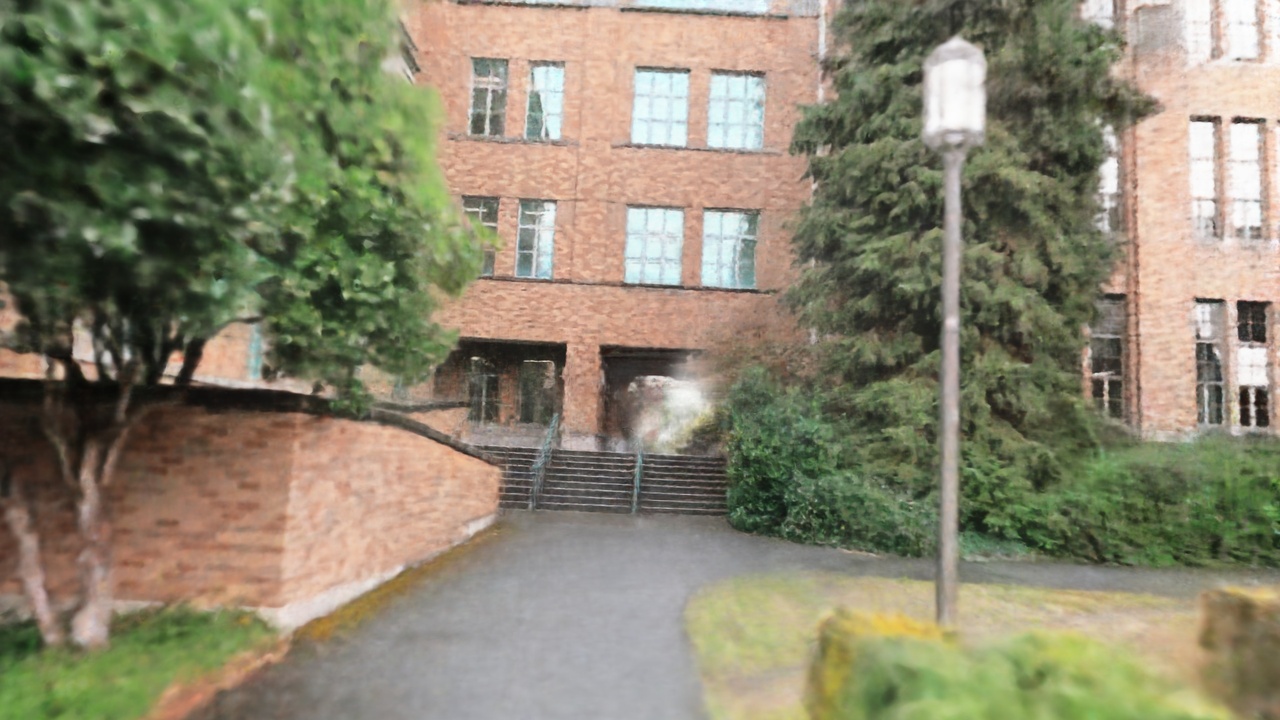} \\
(a) All at once & (b) Progressive \\
\end{tabular}%
\vspace{-2mm}
\caption{\textbf{Importance of progressive optimization.} (a) When estimating all camera poses at once and from scratch, the relationship between the poses is lost and we obtain this disjointed camera trajectory. In this scene, the camera follows a forward trajectory, coherent with (b) the path estimated with progressive optimization.}
\label{fig:prog_impact}
\vspace{-4mm}
\end{figure}

%% file: figure/fig_loc_impact.tex
\begin{figure}[t]
\centering
\small
\setlength{\tabcolsep}{1pt}
\renewcommand{\arraystretch}{0.6}
\begin{tabular}{cc}
\centeredtab{\includegraphics[width=0.49\columnwidth]{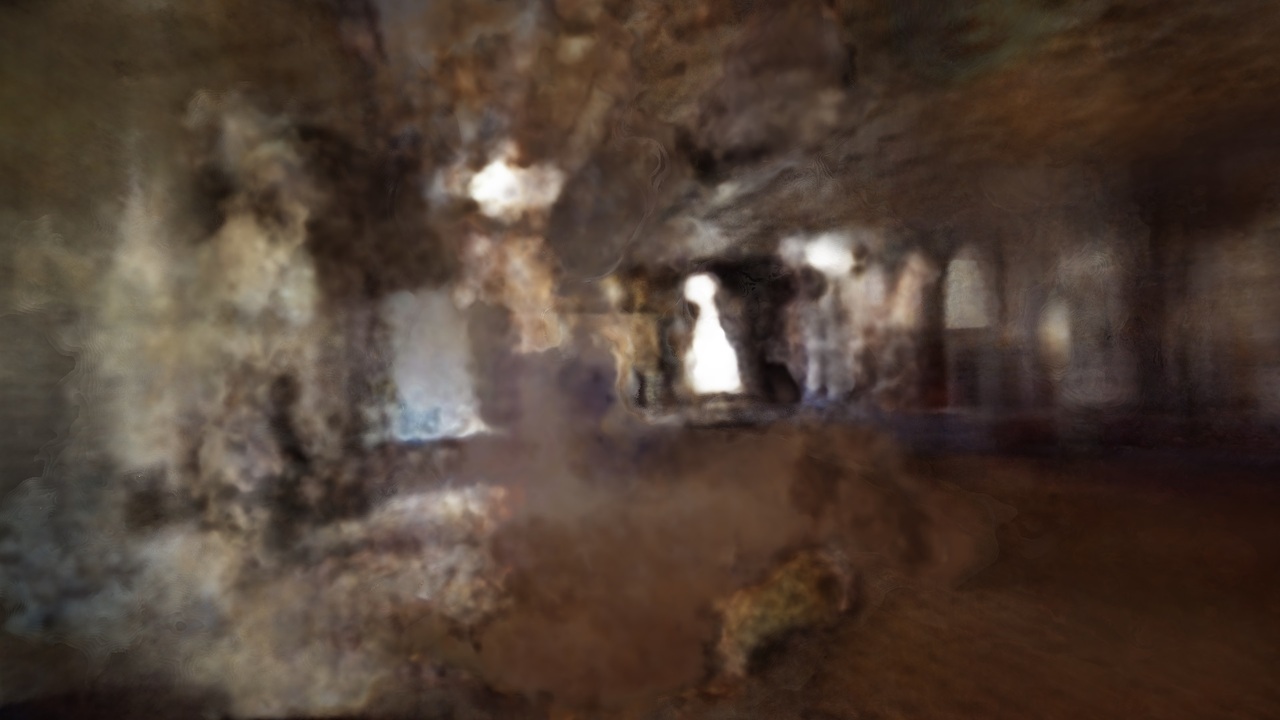}} &
\centeredtab{\includegraphics[width=0.49\columnwidth]{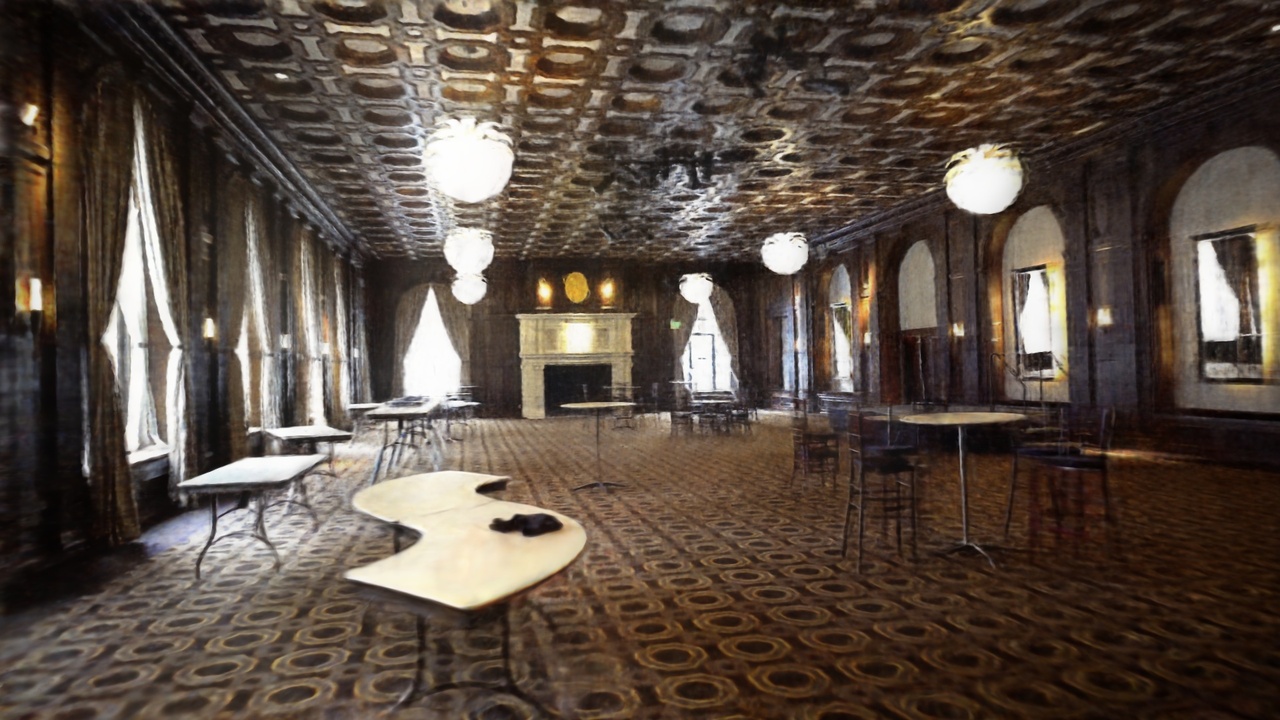}} \\
\centeredtab{\includegraphics[width=0.49\columnwidth]{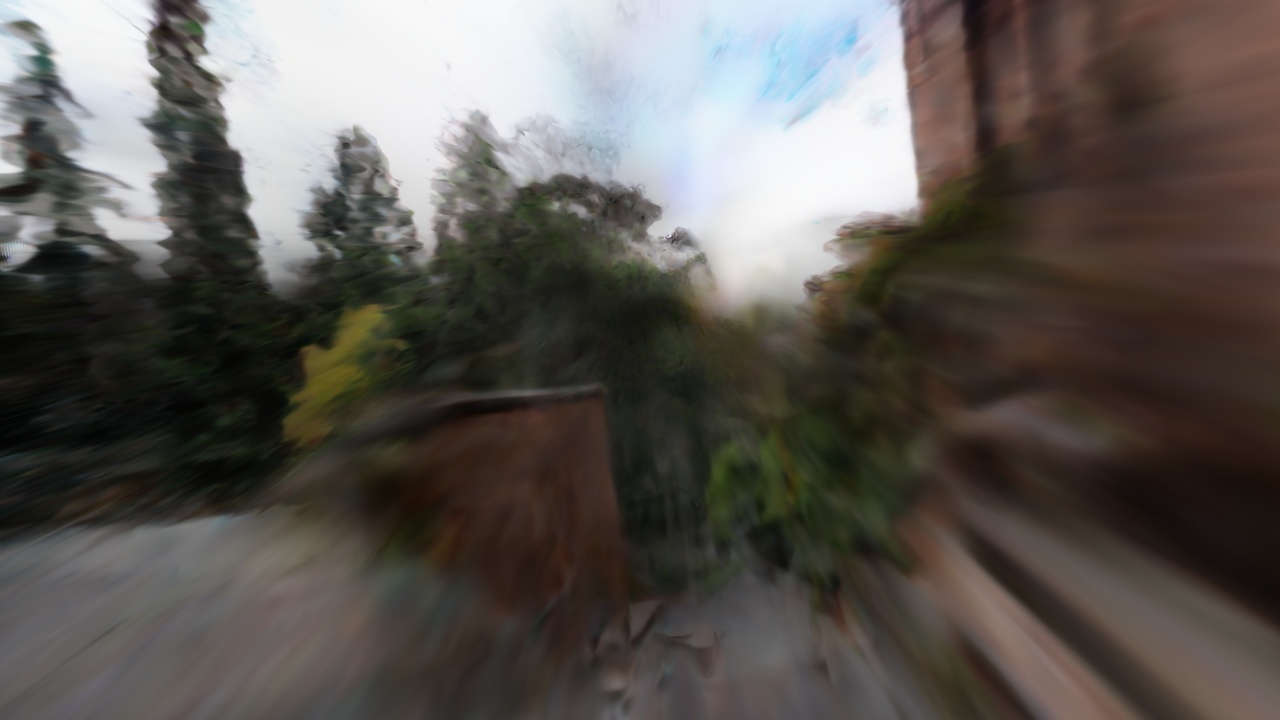}} &
\centeredtab{\includegraphics[width=0.49\columnwidth]{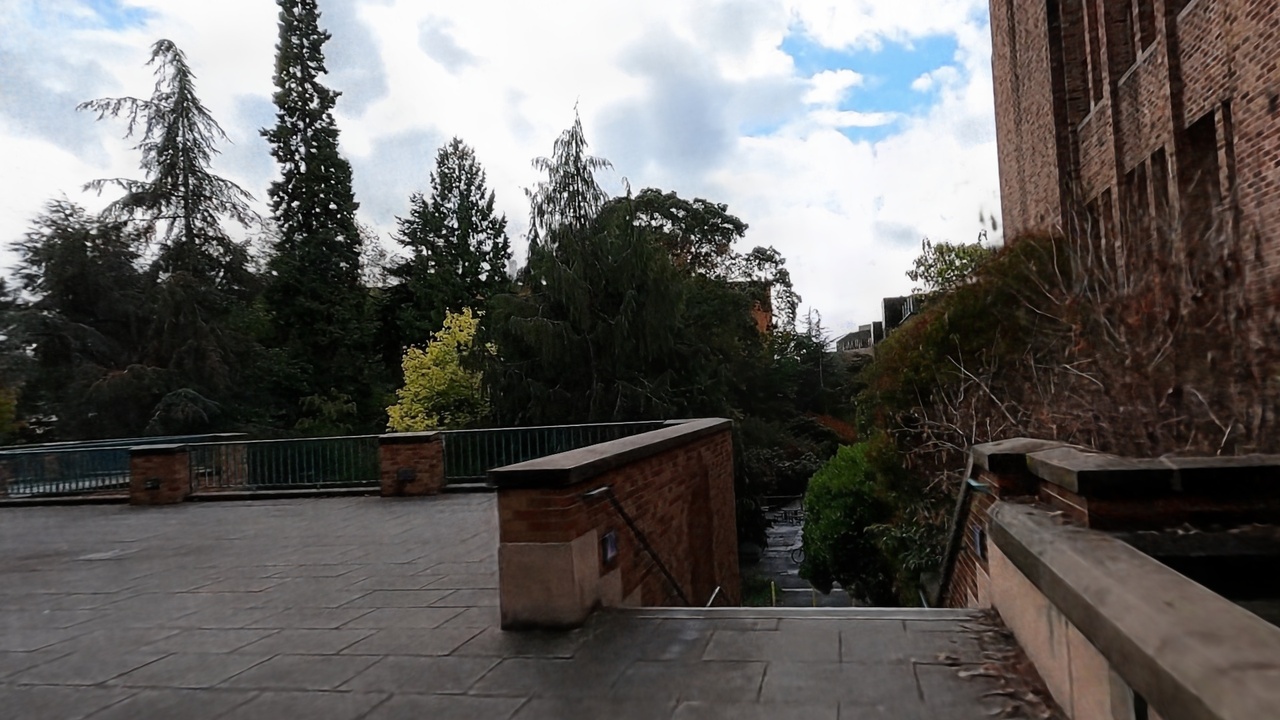}} \\
Global & Local \\
\end{tabular}%
\vspace{-2mm}
\caption{\textbf{Importance of locality.} When using a single global radiance field, the whole scene reconstruction fails if a few camera poses are not estimated correctly (first row). Local radiance fields also allow us to dynamically allocate the representation's capacity around the camera trajectory, producing sharper results (second row). }
\vspace{-4mm}
\label{fig:loc_impact}
\end{figure}

%% file: 4_result.tex
\section{Experimental Results}
\label{sec:result}
\input{figure/fig_ours_graphs}
\input{table/novel_view_synthesis.tex}
\input{figure/fig_tnt_results}

\input{figure/fig_ours_results.tex}

\subsection{Datasets}
\label{sec:dataset}
\topic{Tanks and Temples.} 
We evaluate our method on the \textsc{Tanks and Temples} dataset \cite{Knapitsch2017tankandtemples}. 
We select the sequences without dynamic elements (9 scenes out of 21), retain one in every five frames as motion is slow, downscale the video to full HD resolution (2048$\times$1080 or 1920$\times$1080), and keep the first 1000 images so methods with a static data loader can preload images and rays in a reasonable amount of system memory.

\topic{Static Hikes.} 
We also collect a new dataset with hiking sequences. 
It contains hand-held sequences with larger camera trajectories to test scalability and pose estimation robustness.
It comprises twelve 1920$\times$1080 videos of static outdoor scenes captured with GoPro Hero10 with linear FoV, GoPro Hero9 with narrow FoV, and the wide cameras of LG V60 ThinQ and Samsung Galaxy S21. 

\subsection{Compared Methods} 
\noindent
We compare our method against Mip-NeRF360~\cite{barron2022mipnerf360} and NeRF++~\cite{kaizhang2020nerfpp} as they are both designed to handle unbounded scenes and are suitable for outdoor scenes. For Mip-NeRF360, we set near to 0.1 as we observed clipping with the default value. Nerfacto~\cite{nerfstudio} combines Instant-NGP~\cite{mueller2022instant}'s hash encoding and Mip-NeRF360's scene contraction to efficiently represent unbounded scenes. 
When preprocessed poses are required, we use MultiNeRF~\cite{multinerf2022}'s script to run COLMAP.
We also compare against the scalable representation Mega-NeRF~\cite{Turki_2022_meganerf} (using a 2$\times$2 grid size for our experiments).
Since those methods require pre-processed poses, we include SCNeRF~\cite{SCNeRF2021} and BARF~\cite{lin2021barf} for self-calibrated experiments. 
For SCNeRF, we use the NeRF++ codebase to better represent unbounded scenes. 
Note that SCNeRF requires COLMAP as initialization: it fails in our experiments when optimizing poses from scratch with both the NeRF and NeRF++ bases (we get NaN renders). 
We, therefore, had to exclude SCNeRF from the fully self-calibrated evaluations. 

\subsection{Quantitative Evaluation}
\label{sec:quanitative}
\noindent
To quantitatively evaluate the synthesized novel views, we select every ten frames as a \emph{test} image. 
We show the PSNR, SSIM, and LPIPS~\cite{zhang2018perceptual} between the synthesized views and the corresponding ground truth views in \tabref{tnt}. 
Averages are computed in the square error domain for PSNR and $\sqrt{1 - \text{SSIM}}$ domain for SSIM, following~\cite{Brunet2012ssimmaths}.
For the self-calibrated experiments, we estimate the test poses by adding iterations that only optimize the poses, \emph{without} updating the intrinsic or the radiance fields' parameters. 

Table~\ref{tab:tnt} shows that, although the videos are not what we target, with a smaller camera path and several inward-looking 360 scenes, our method provides competitive results. 
With COLMAP poses, we obtain similar quality as Mip-NeRF360~\cite{barron2022mipnerf360}.
Mega-NeRF~\cite{Turki_2022_meganerf}, being designed for a different type of input data, shows lower quality despite using several radiance fields.
Compared to other self-calibrating radiance fields methods~\cite{SCNeRF2021,lin2021barf}, we obtain much better results when optimizing poses from scratch thanks to our progressive optimization that allows for fewer parameters to be estimated from scratch at once and adds a flexible locality prior to the camera pose. 

Figure~\ref{fig:ours_graphs} shows that, on the \textsc{Static Hikes} dataset featuring longer trajectories and more challenging scenes, we consistently obtain better results than the self-calibrated method BARF~\cite{lin2021barf}. 
Mip-NeRF360~\cite{barron2022mipnerf360}, relying on COLMAP, cannot produce results for $19.5\%$ of the test frames. 
In addition, we show higher quality on the rendered frames.
\input{figure/extrapolation_main}

\subsection{Qualitative evaluation}
\label{sec:visual_comparison}
\noindent 
Figure~\ref{fig:tnt_results} compares the results on the \textsc{Tanks and Temples} dataset, and Figures~\ref{fig:ours_results} and \ref{fig:extra} show results on the \textsc{Static Hikes} dataset we acquired. 
Our approach can provide results for all frames and maintain better sharpness throughout the entire trajectory.

\subsection{Ablation Study}
\label{sec:ablation}
\input{table/ablation.tex}
\noindent
Table~\ref{tab:ablation} shows that our progressive optimization and local radiance fields are both necessary to obtain our results. 
Furthermore, Figure~\ref{fig:prog_impact} highlights that progressive optimization is crucial when estimating poses for long sequences, and Figure~\ref{fig:loc_impact} shows that local radiance fields grant more robustness and allow for scalability. 

%% file: figure/fig_ours_graphs.tex
\begin{figure*}[th]
\centering
\small
\setlength{\tabcolsep}{0pt}
\renewcommand{\arraystretch}{0.5}
\begin{tabular}{cccccc}
\centeredtab{\rotatebox[origin=c]{90}{PSNR $\uparrow$}} &
\centeredtab{\includegraphics[trim={8mm 1mm 0mm 7mm},clip,width=0.65\columnwidth]{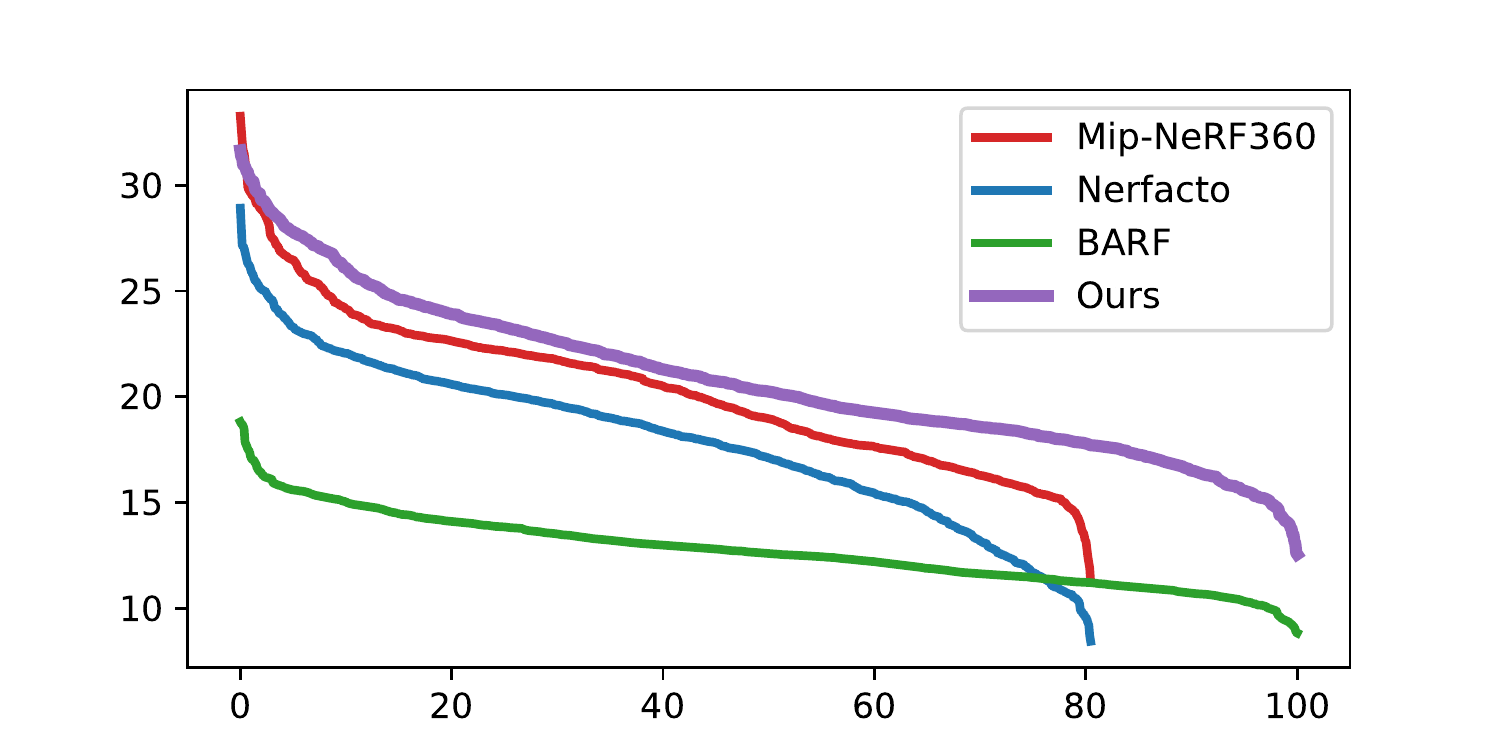}} &
\centeredtab{\rotatebox[origin=c]{90}{SSIM $\uparrow$}} &
\centeredtab{\includegraphics[trim={8mm 1mm 0mm 7mm},clip,width=0.65\columnwidth]{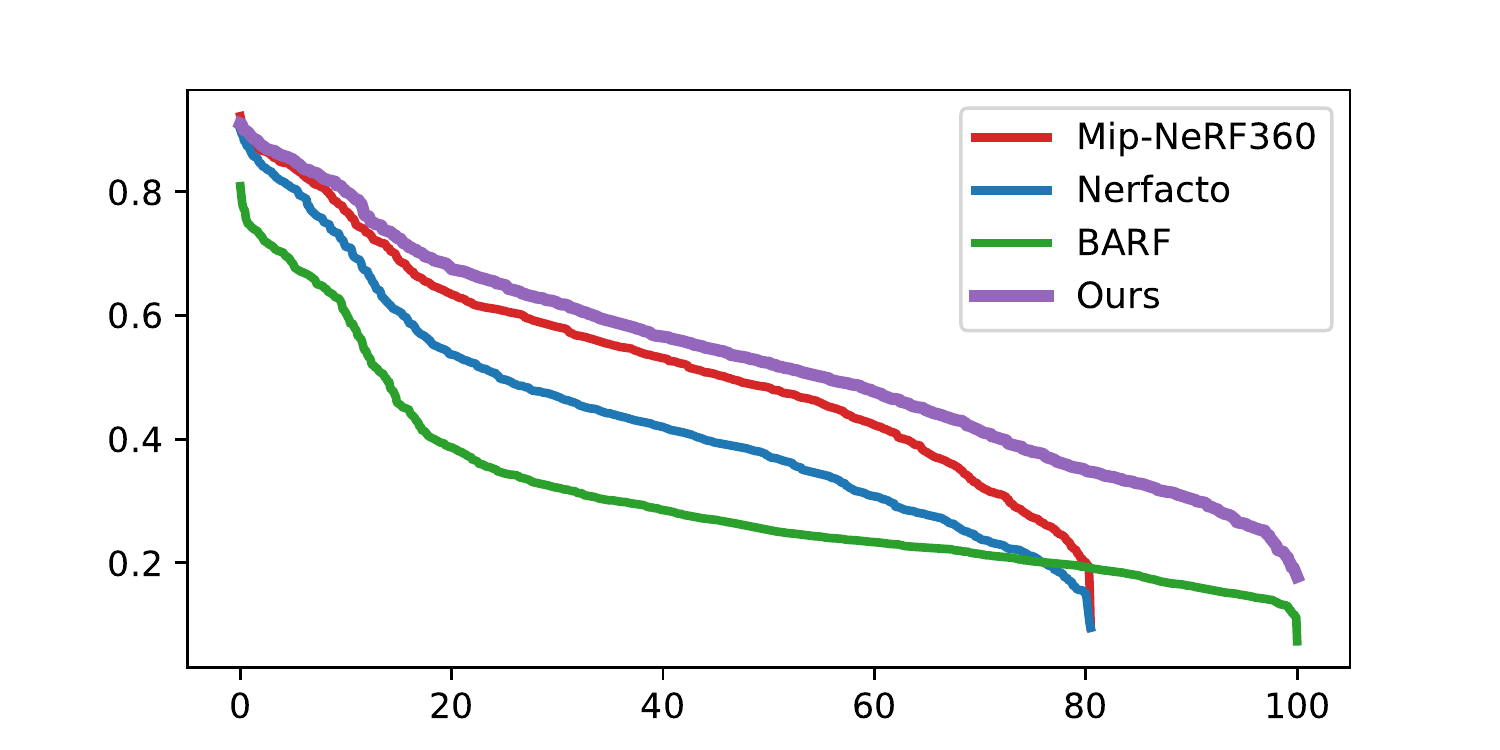}} &
\centeredtab{\rotatebox[origin=c]{90}{LPIPS $\downarrow$}} &
\centeredtab{\includegraphics[trim={8mm 1mm 0mm 7mm},clip,width=0.65\columnwidth]{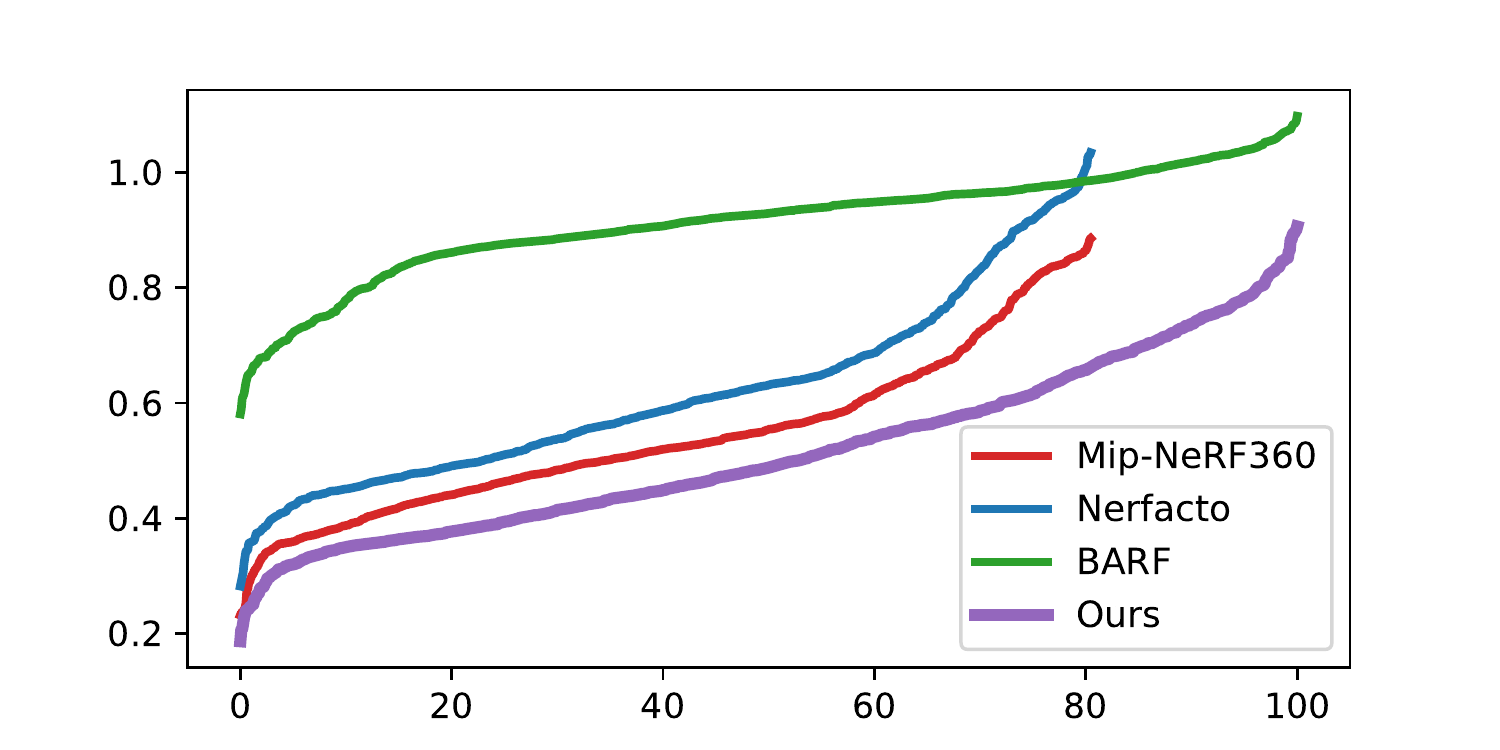}} \\
 & \% test frame & & \% test frame & & \% test frame
\end{tabular}%
\vspace{-2mm}
\caption{\textbf{Novel view synthesis evaluation on the \textsc{Static Hikes} dataset.} We plot the sorted frame-wise metrics to assess the quality distribution across the dataset. Since COLMAP is not able to register poses for all frames, Mip-NeRF360's and Nerfacto's curves terminate early. Only BARF and our method estimate pose for each frame, but BARF cannot maintain high accuracy when optimizing a large number of camera poses at once.}
\label{fig:ours_graphs}
\vspace{-2mm}
\end{figure*}

%% file: table/novel_view_synthesis.tex
\begin{table}[t]
\caption{
\textbf{Novel view synthesis results.}
We report the average PSNR, SSIM and LPIPS results with comparisons to existing methods on \textsc{Tank and Temples} dataset. The best performance is in \first{bold} and the second best is \second{underscored}. 
The COLMAP rows use COLMAP camera parameters as initialization. When fixed, we keep the poses throughout the entire optimization. When refined, we jointly optimize poses and radiance fields. The self-calibrated methods start pose optimization from scratch.
}
\label{tab:tnt}
\vspace{-3mm}
\footnotesize
\centering
\begin{tabular}{lll ccc}
\toprule
\multicolumn{2}{c}{\centeredtab{Pose \\ estimation}} & & PSNR $\uparrow$ & SSIM $\uparrow$ & LPIPS $\downarrow$ \\
\midrule
\multirow{7}{*}{\rotatebox[origin=c]{90}{COLMAP}}
&
\multirow{4}{*}{\rotatebox[origin=c]{90}{fixed}}
& NeRF++~\cite{kaizhang2020nerfpp}         & 19.59 & 0.562 & 0.682 \\
& & Mega-NeRF~\cite{Turki_2022_meganerf}     & 18.09 & 0.548 & 0.622 \\
& & Nerfacto~\cite{nerfstudio}  & 16.85 & 0.641 & \second{0.466} \\
& &  Mip-NeRF360~\cite{barron2022mipnerf360} & \second{21.09} & \first{0.714} & \first{0.406} \\
& & LocalRF (ours)                                     & \first{22.40} & \second{0.667} & 0.494 \\
\cmidrule{2-6}
&
\multirow{3}{*}{\rotatebox[origin=c]{90}{refined}}
& SCNeRF~\cite{SCNeRF2021}                   & \second{15.88} & \second{0.506} & \second{0.708}  \\
& & BARF~\cite{lin2021barf}                  &  9.62 & 0.383 & 0.893  \\
& & LocalRF (ours)                                     & \first{22.85} & \first{0.676} & \first{0.475}  \\
\midrule
\multirow{2}{*}{\rotatebox[origin=c]{90}{Self} \rotatebox[origin=c]{90}{calib.}}
& & BARF~\cite{lin2021barf}                  & \second{11.78} & \second{0.430} & \second{0.884}  \\
& & LocalRF (ours)                                     & \first{20.41} & \first{0.593} & \first{0.624}  \\
\bottomrule
\end{tabular}
\end{table}

%% file: figure/fig_tnt_results.tex
\begin{figure*}[t]
\small
\centering
\setlength{\tabcolsep}{1pt}
\renewcommand{\arraystretch}{0.6}
\begin{tabular}{ccccc}
\centeredtab{(a)} & 
\centeredtab{\includegraphics[width=0.235\textwidth]{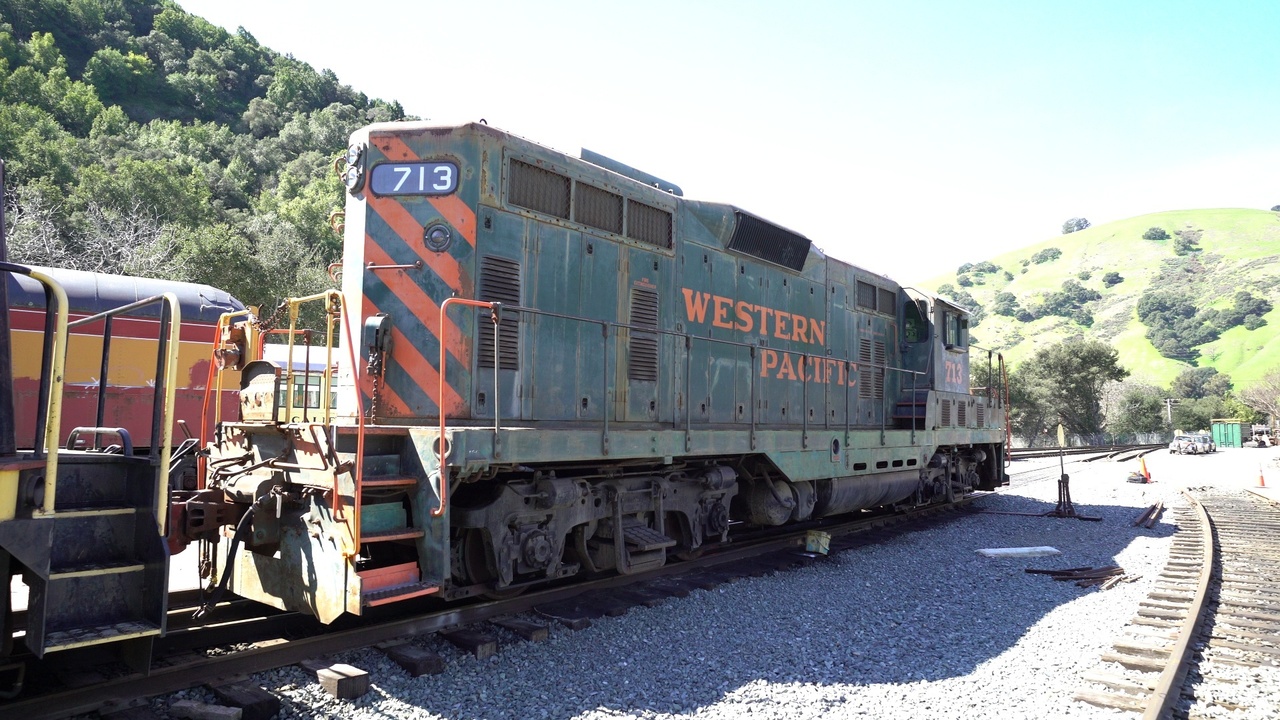}} &
\centeredtab{\includegraphics[width=0.235\textwidth]{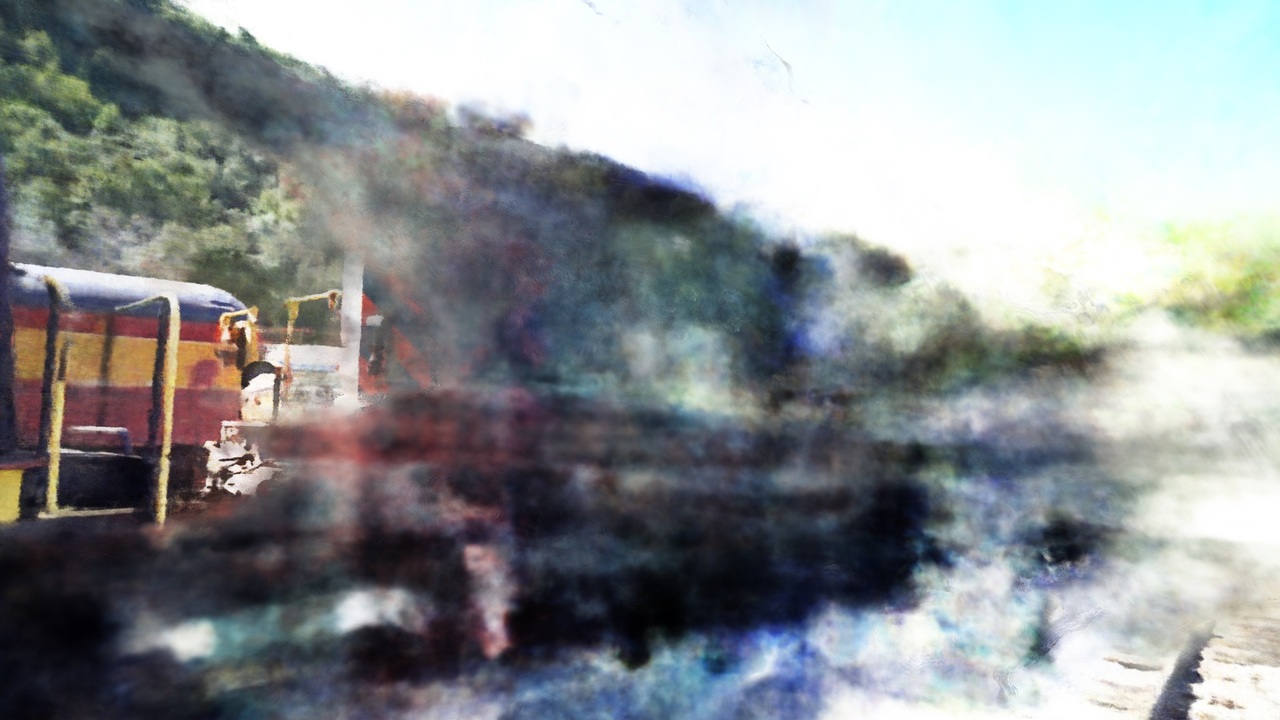}} &
\centeredtab{\includegraphics[width=0.235\textwidth]{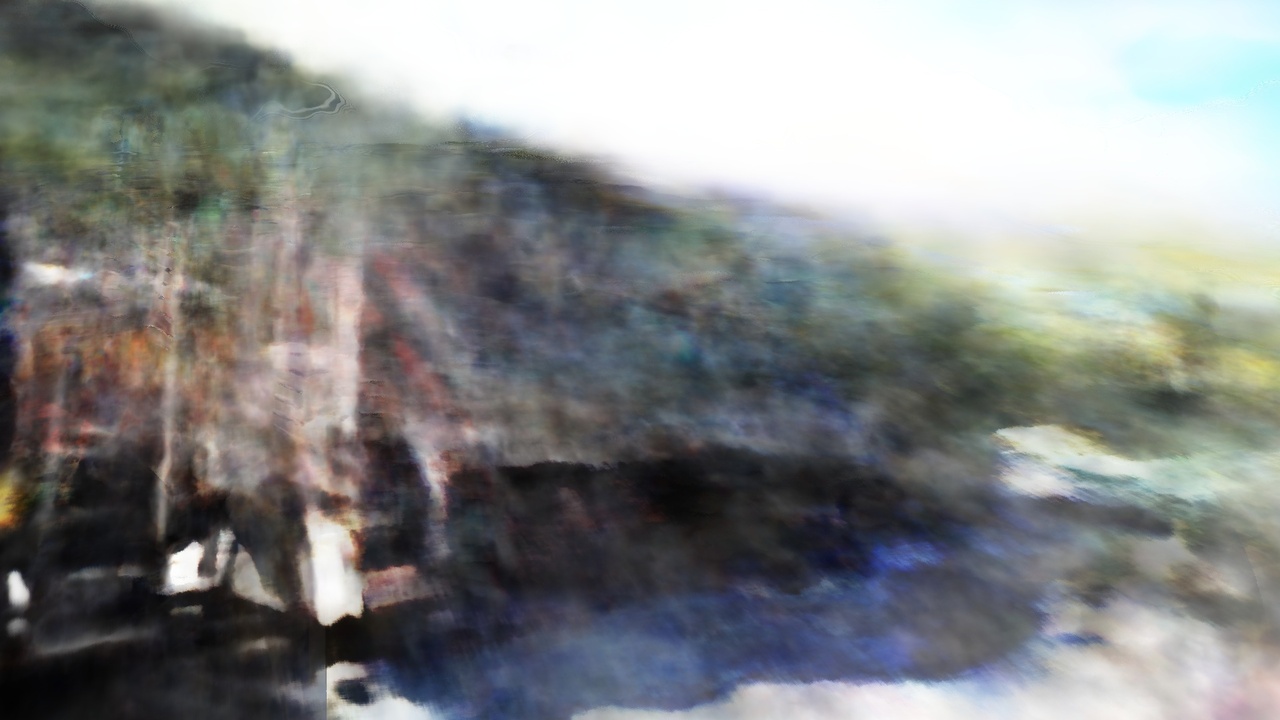}} &
\centeredtab{\includegraphics[width=0.235\textwidth]{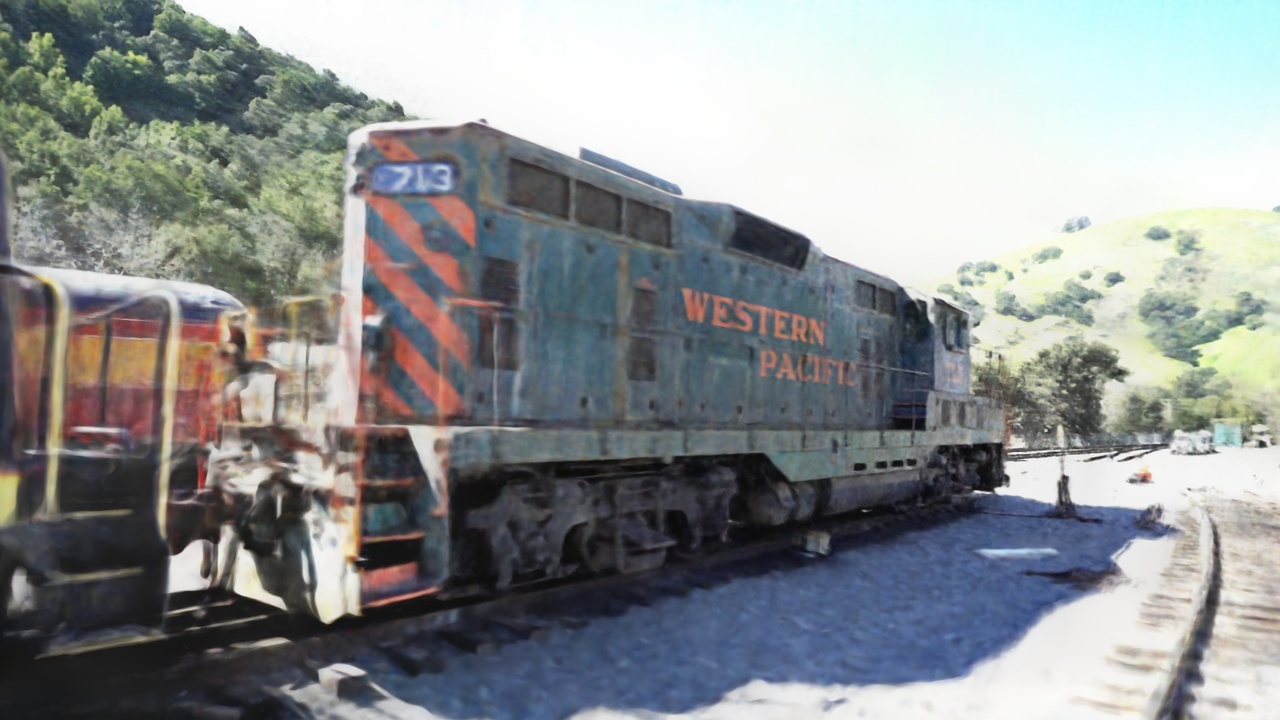}} \\
\centeredtab{(b)} &
\centeredtab{\includegraphics[width=0.235\textwidth]{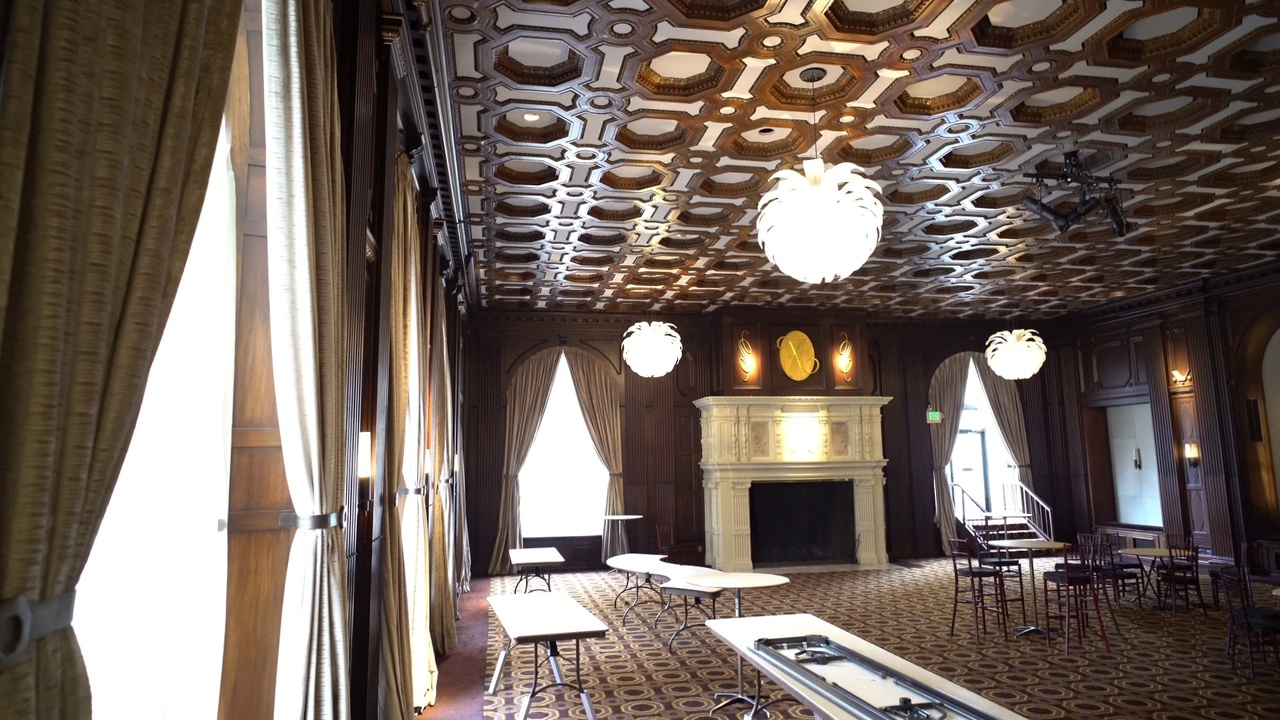}} &
\centeredtab{\includegraphics[width=0.235\textwidth]{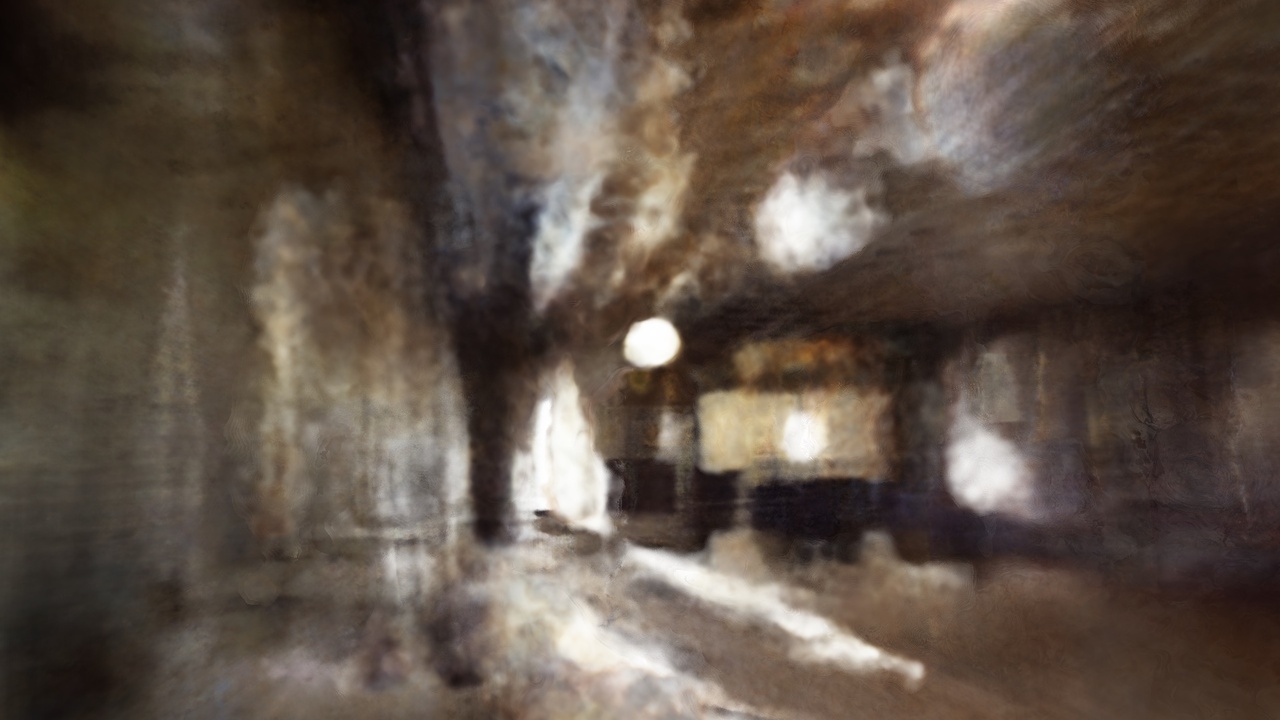}} &
\centeredtab{\includegraphics[width=0.235\textwidth]{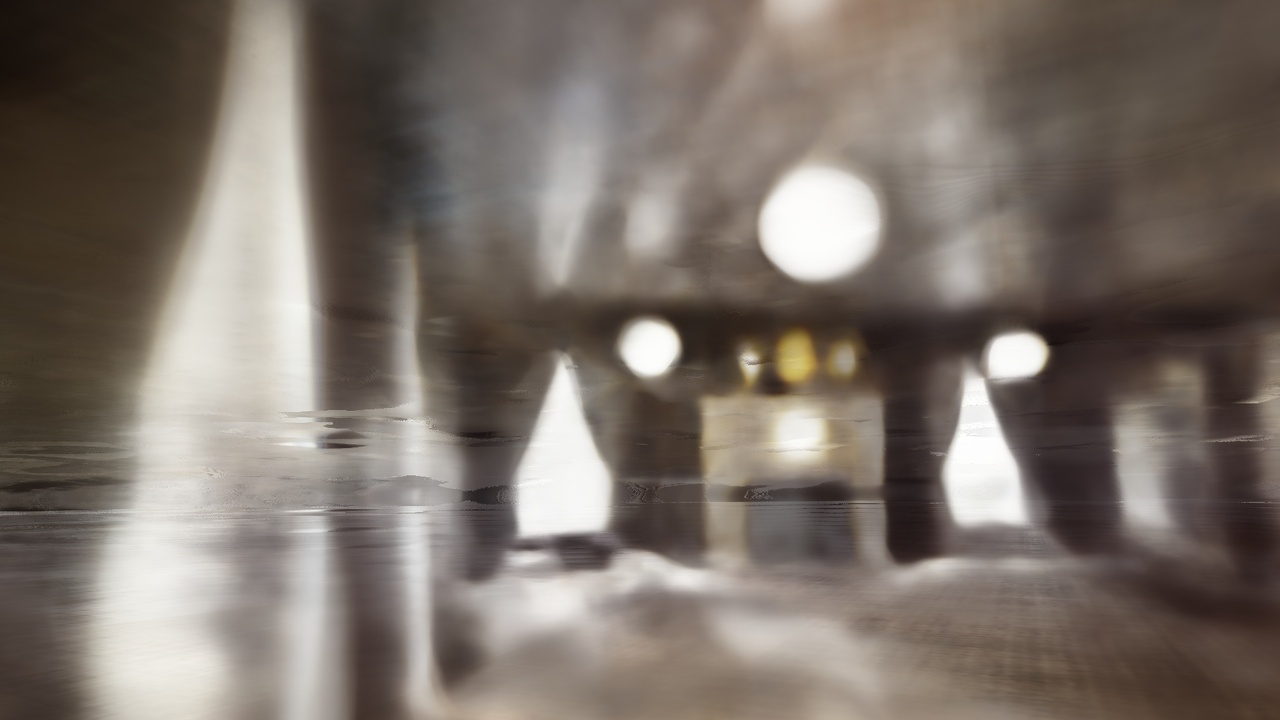}} &
\centeredtab{\includegraphics[width=0.235\textwidth]{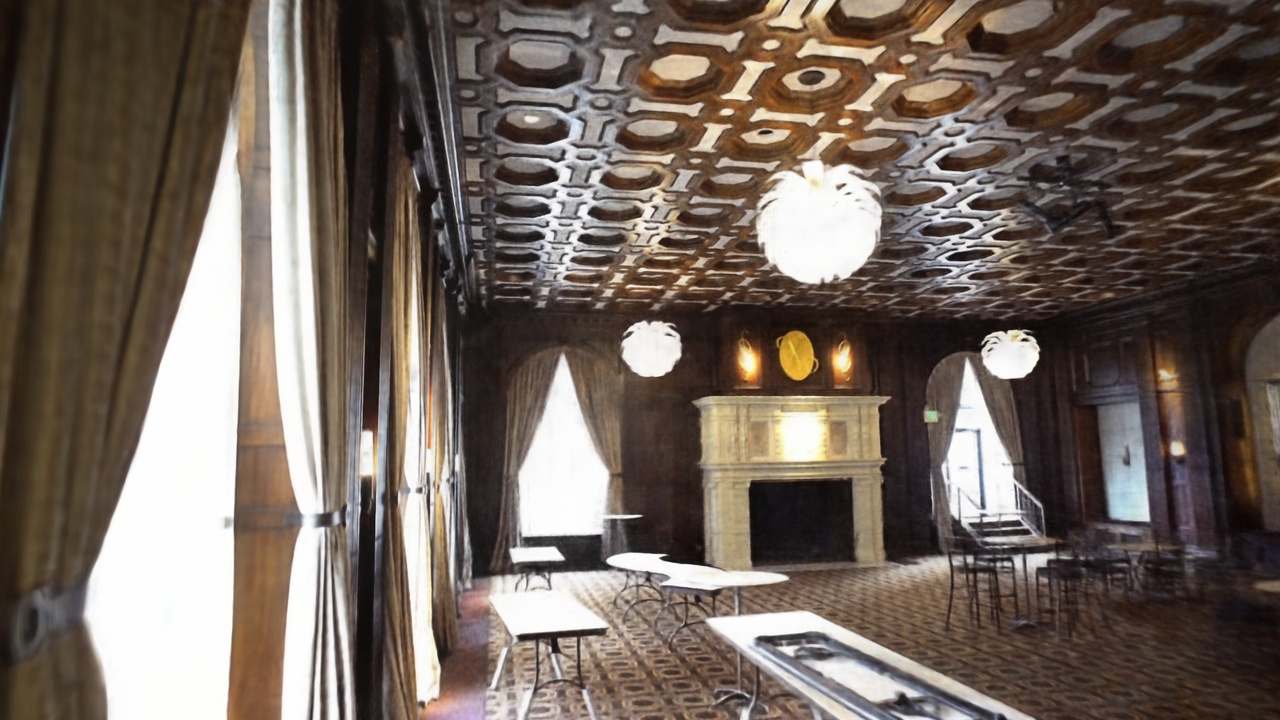}} \\
\centeredtab{(c)} &
\centeredtab{\includegraphics[width=0.235\textwidth]{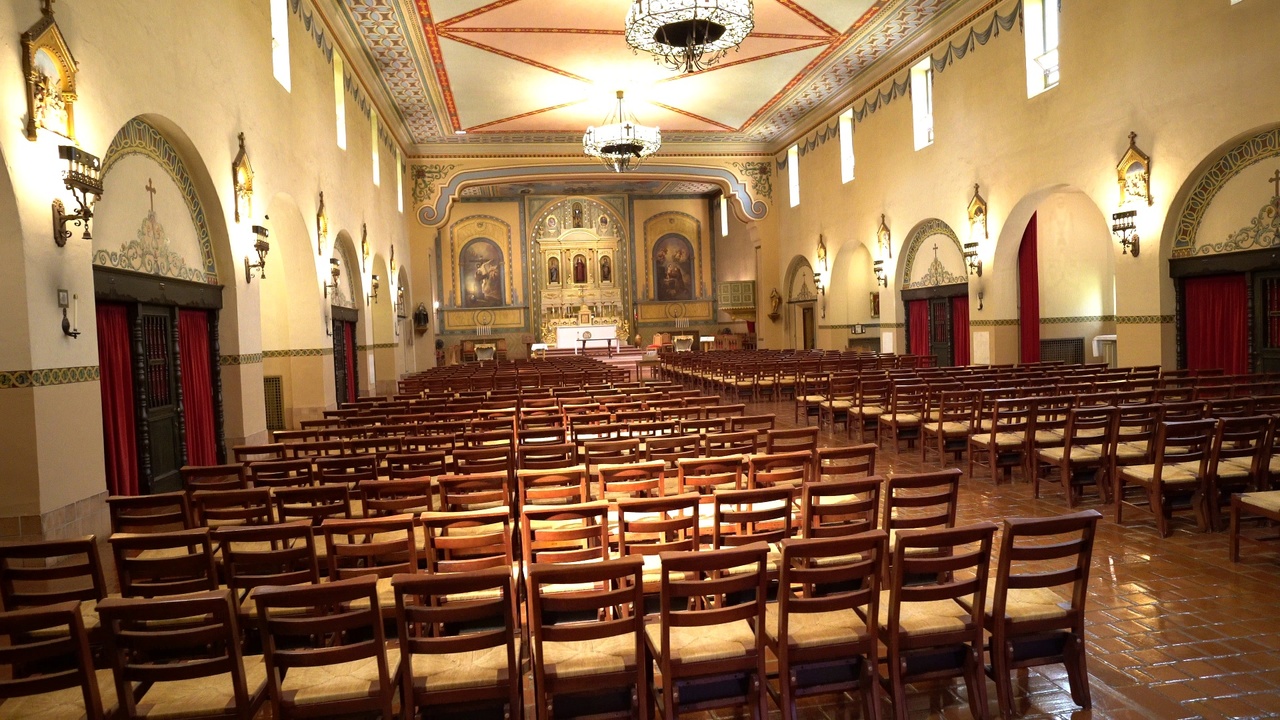}} &
\centeredtab{\includegraphics[width=0.235\textwidth]{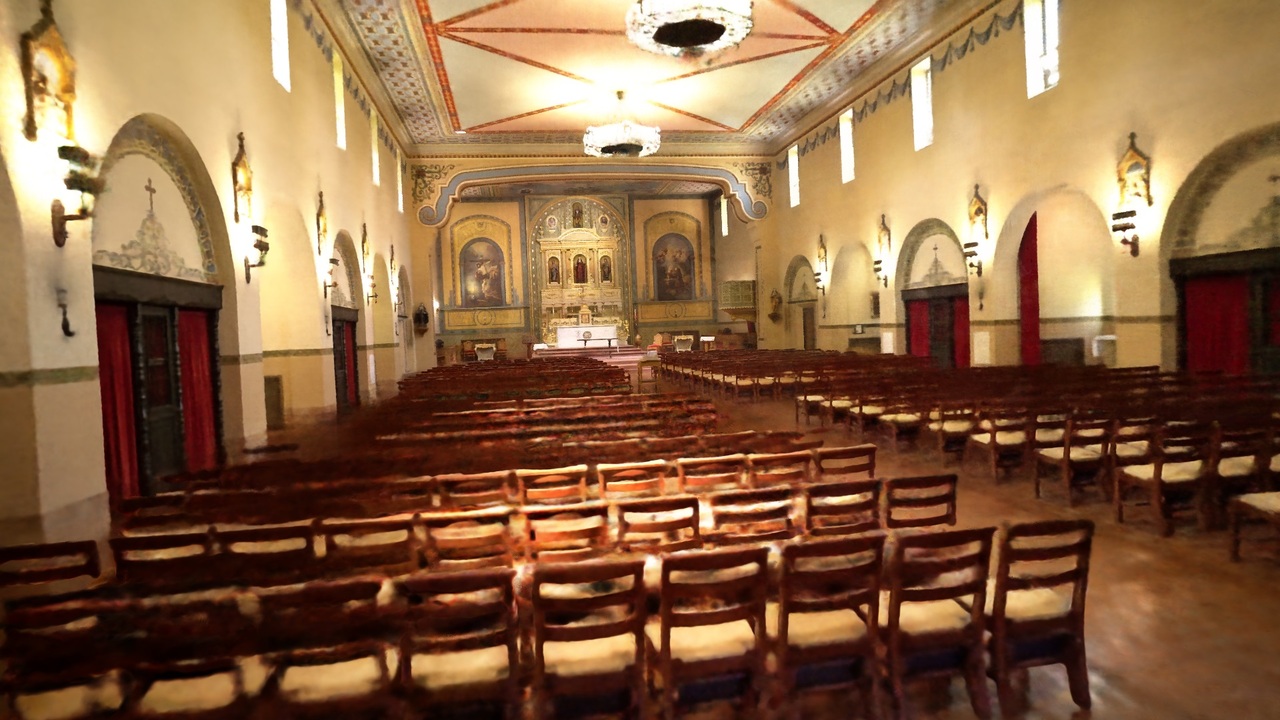}} &
\centeredtab{\includegraphics[width=0.235\textwidth]{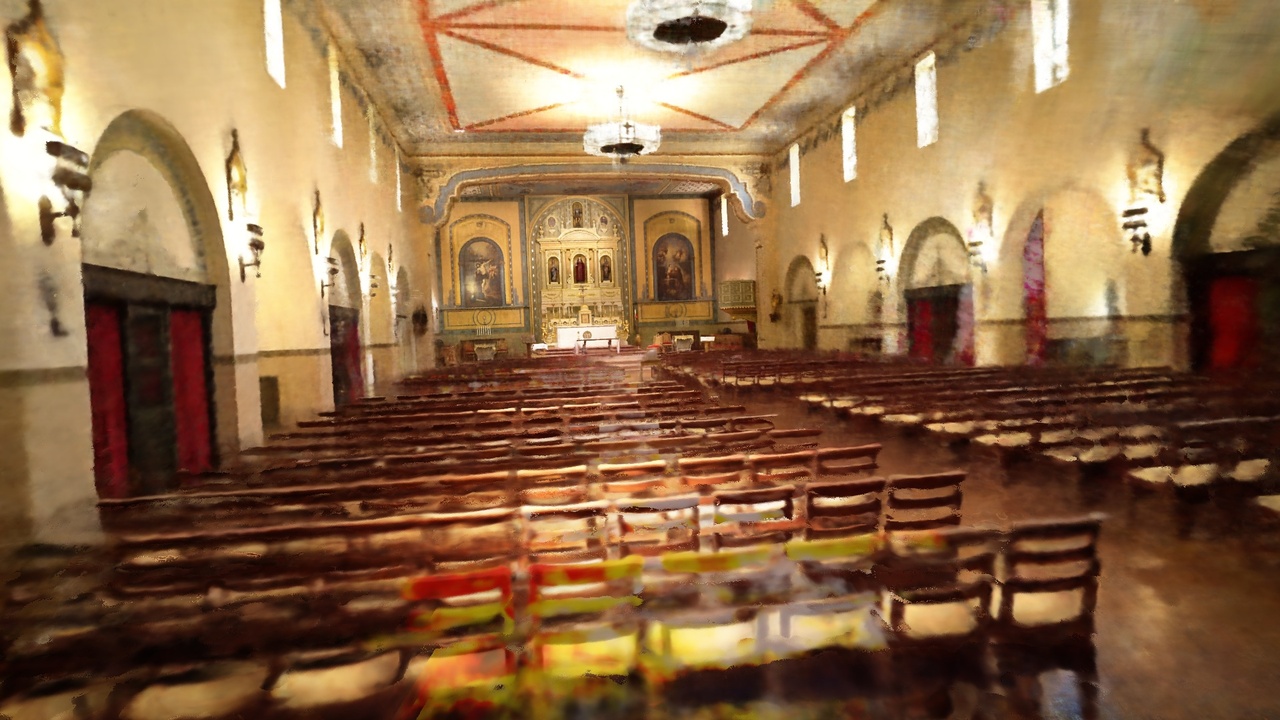}} &
\centeredtab{\includegraphics[width=0.235\textwidth]{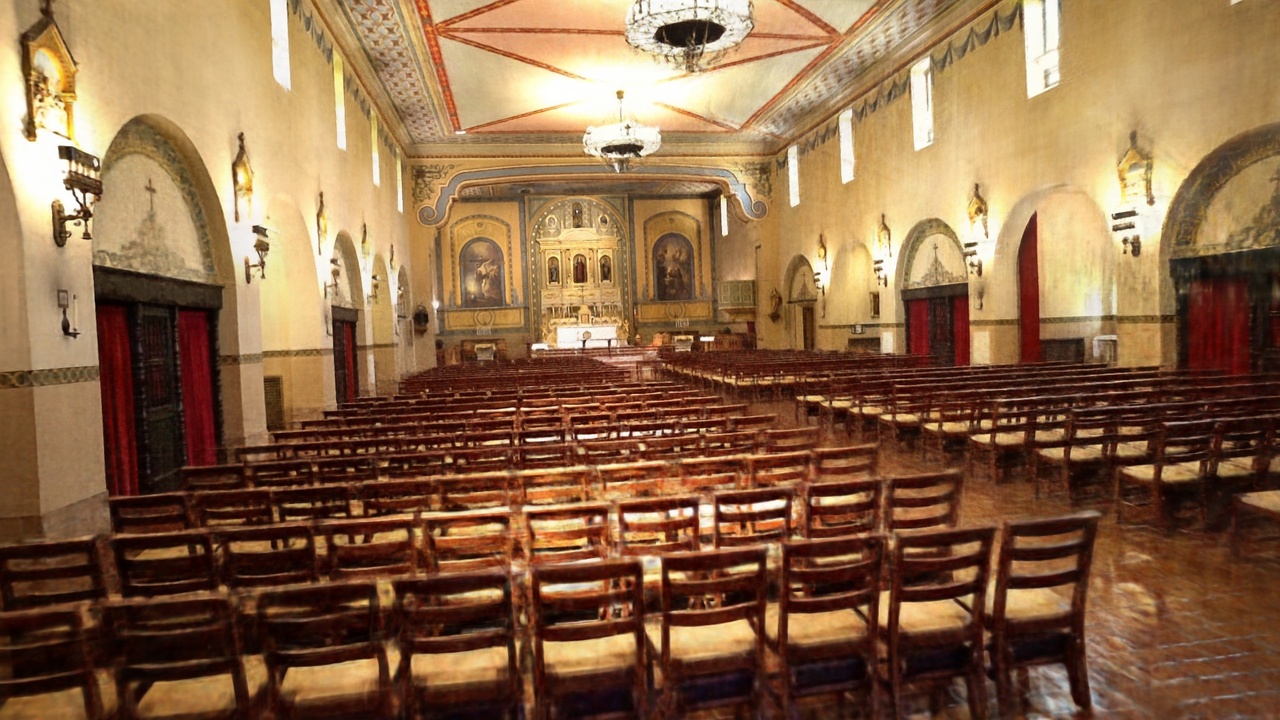}} \\
& Ground truth & Mip-NeRF360 & Mega-NeRF & LocalRF (ours) \\
\end{tabular}%
\vspace{-3mm}
\caption{\textbf{Novel view synthesis results on \textsc{Tanks and Temples} dataset.} (a) and (b) Locality allows for more robustness to illumination changes and pose estimation failures. (c) We can obtain sharper results since the less contracted space follows the trajectory.
}
\label{fig:tnt_results}
\vspace{-3mm}
\end{figure*}

%% file: figure/fig_ours_results.tex
\begin{figure*}[t]
\small
\centering
\setlength{\tabcolsep}{1pt}
\renewcommand{\arraystretch}{0.6}
\begin{tabular}{cccc}
\centeredtab{\includegraphics[width=0.245\textwidth]{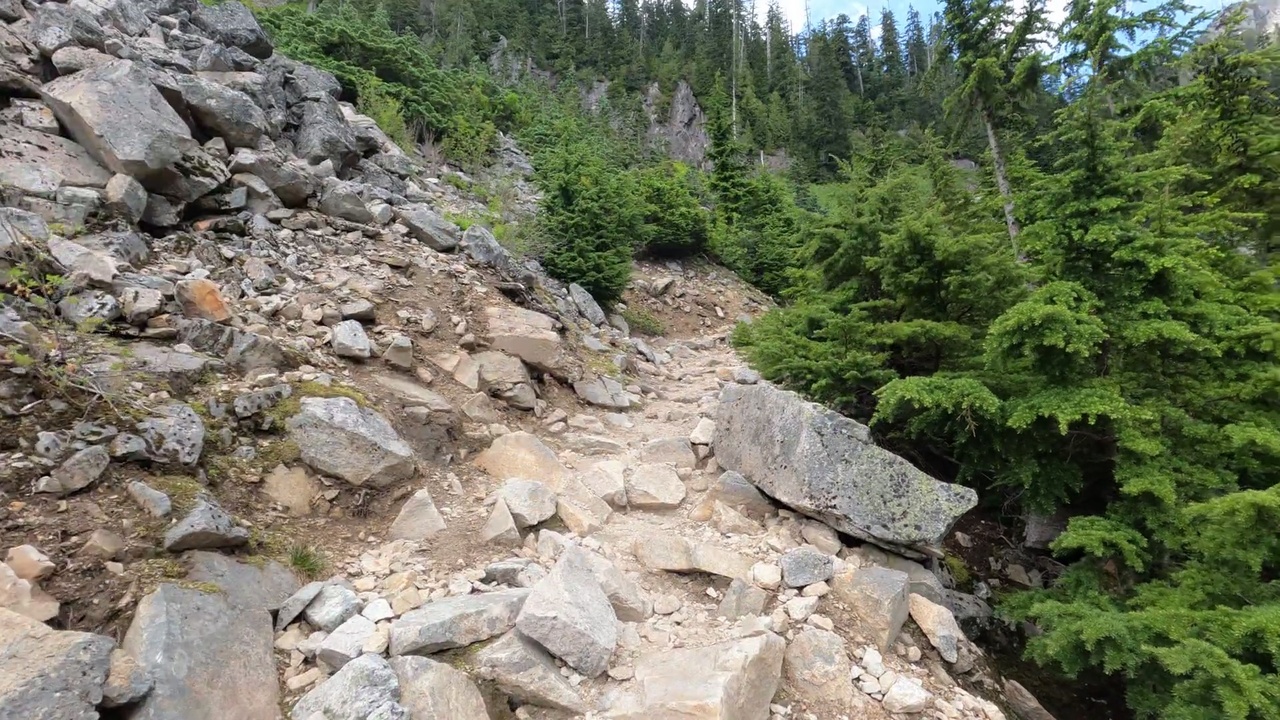}} &
\centeredtab{\includegraphics[width=0.245\textwidth]{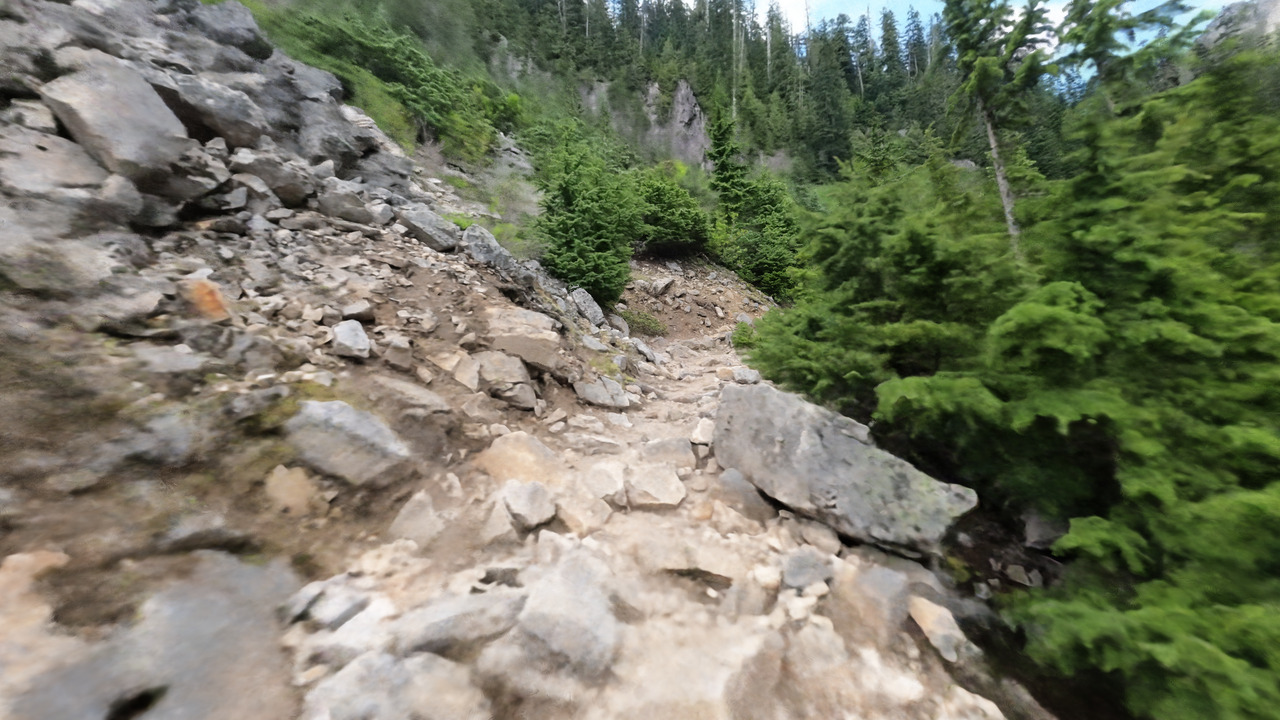}} &
\centeredtab{\includegraphics[width=0.245\textwidth]{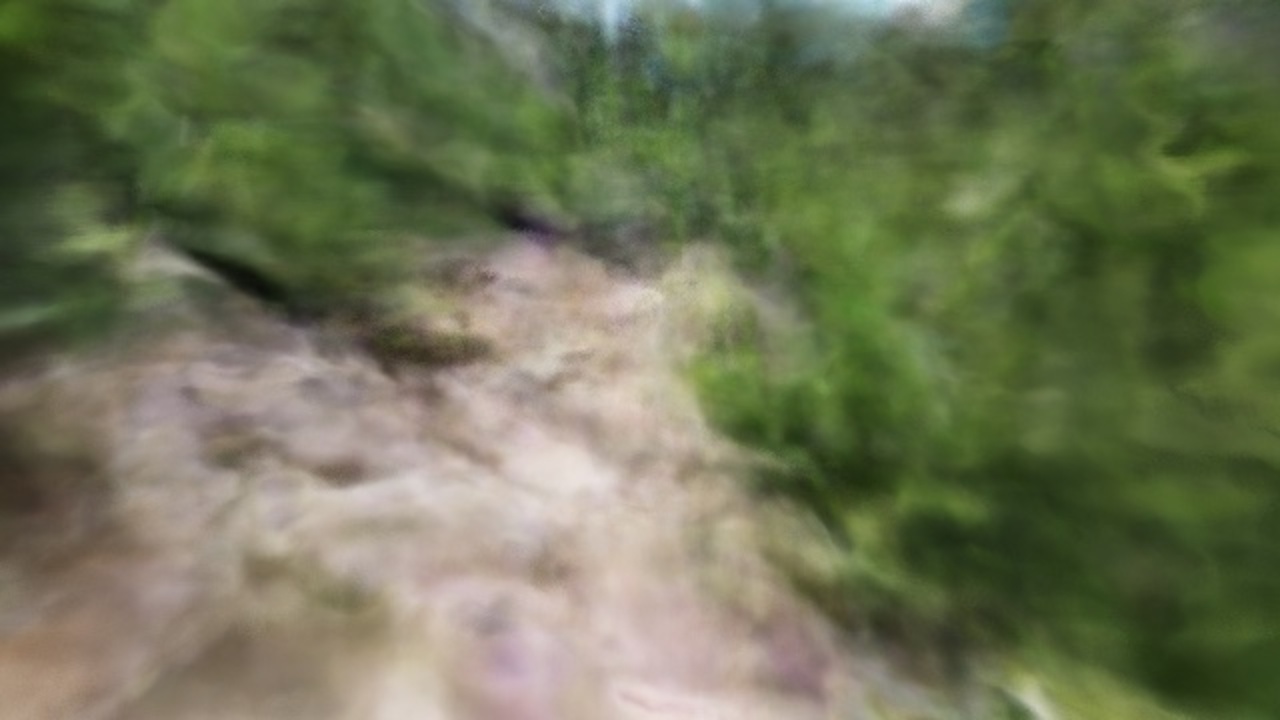}} &
\centeredtab{\includegraphics[width=0.245\textwidth]{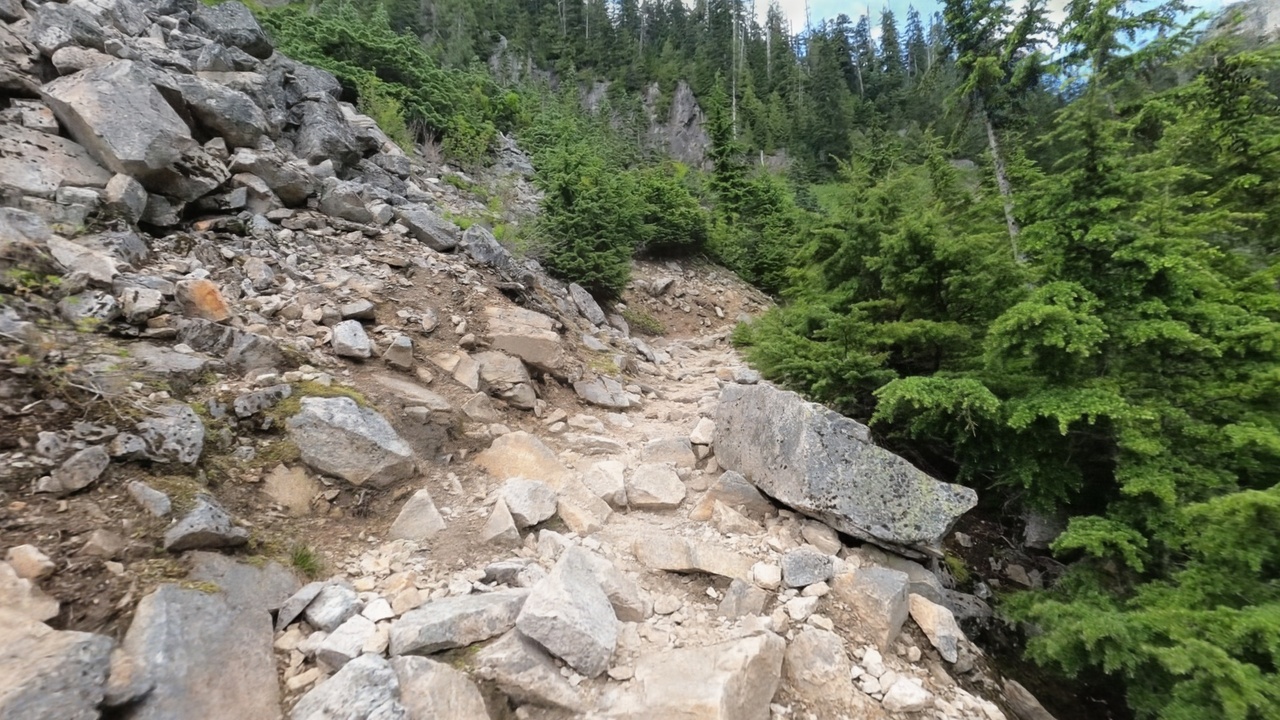}}
\\
\centeredtab{\includegraphics[width=0.245\textwidth]{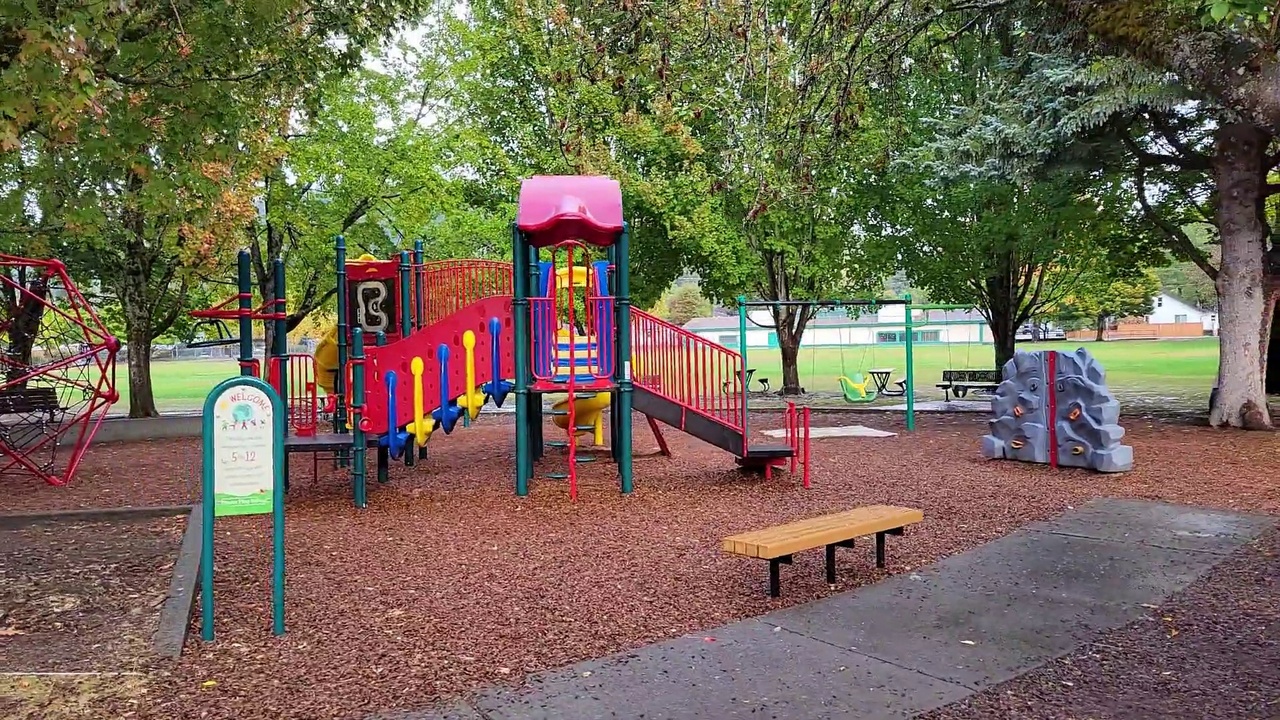}} &
\centeredtab{COLMAP fails} &
\centeredtab{\includegraphics[width=0.245\textwidth]{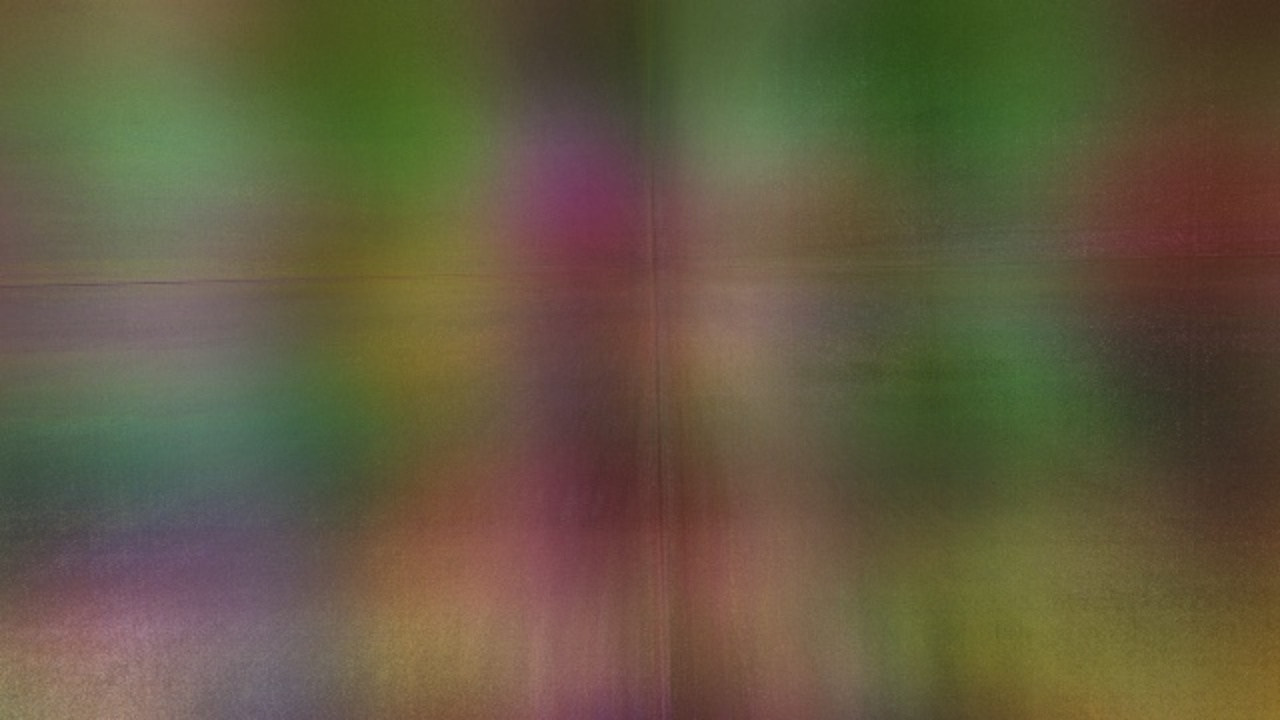}} &
\centeredtab{\includegraphics[width=0.245\textwidth]{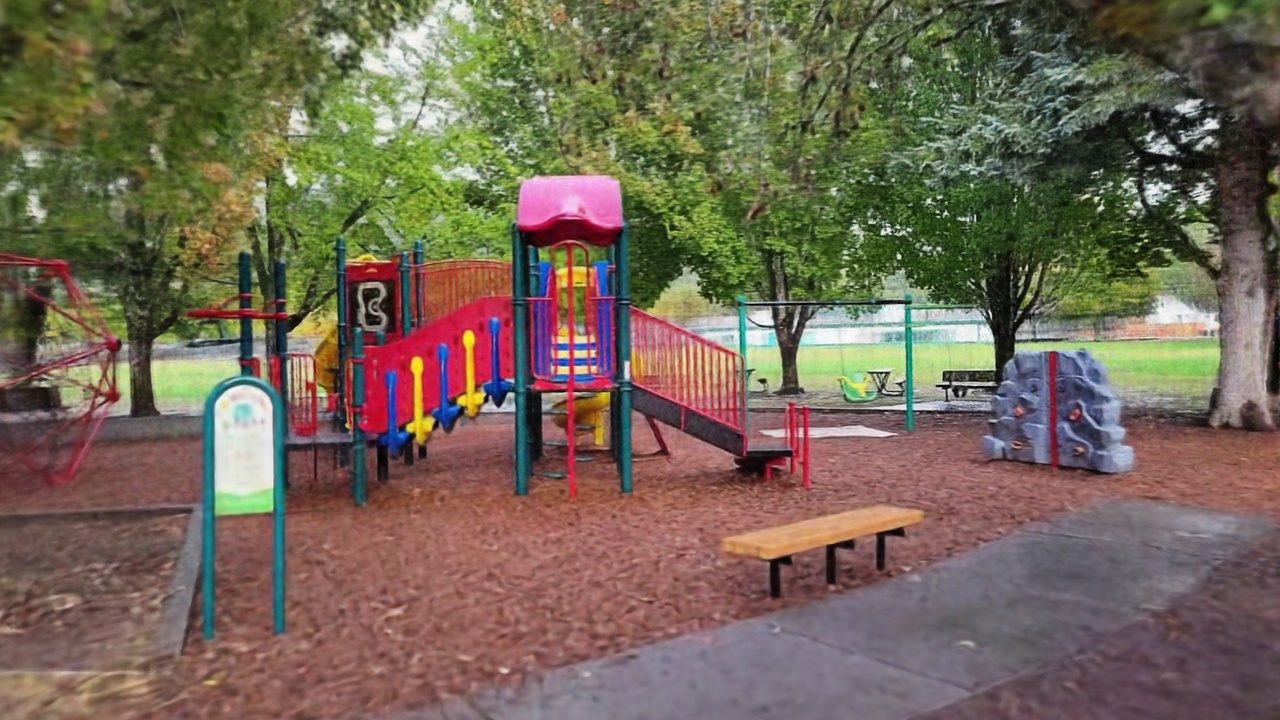}}
\\
\centeredtab{\includegraphics[trim={0 36cm 0 0},clip,width=0.245\textwidth]{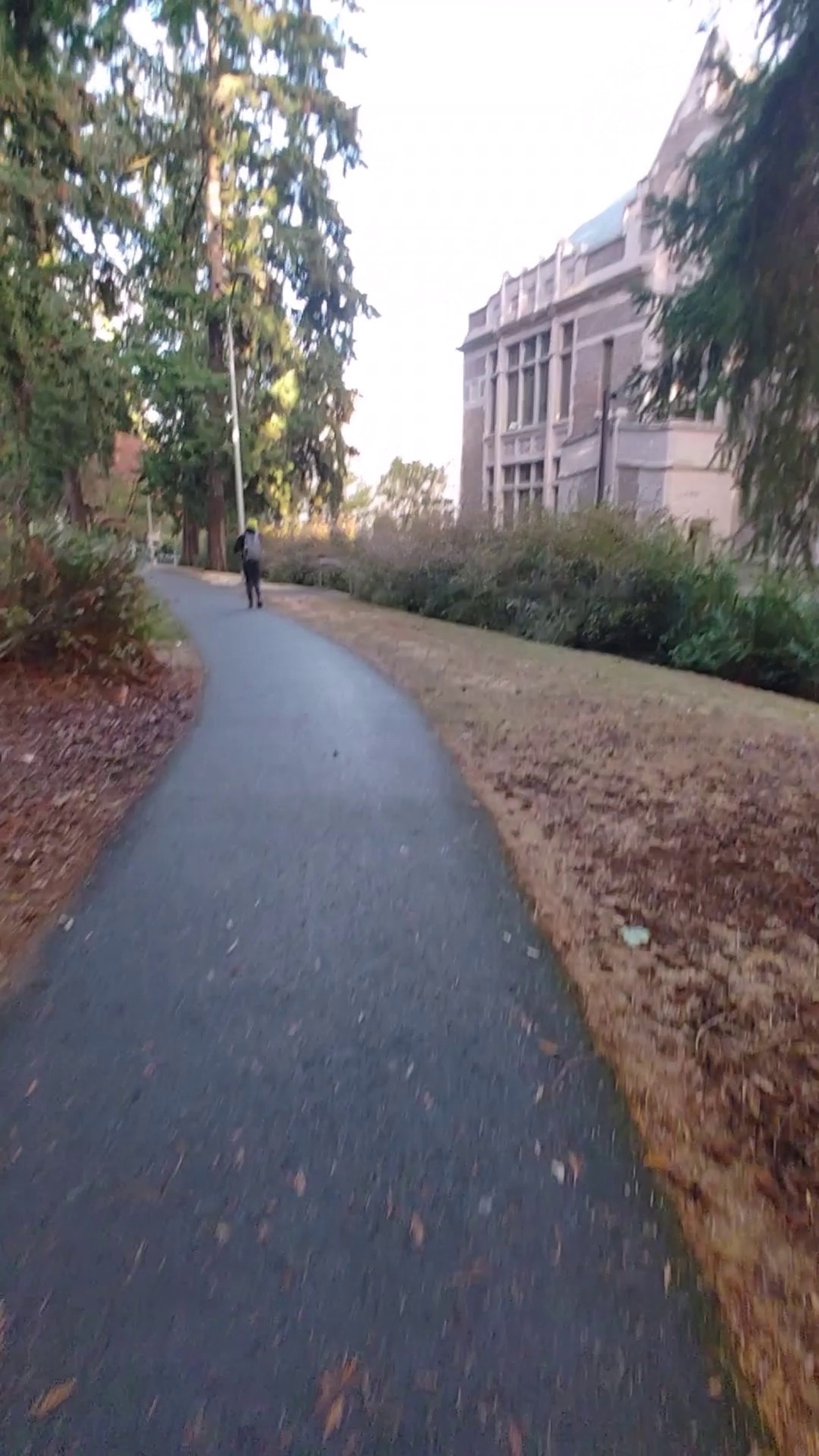}} &
\centeredtab{\includegraphics[trim={0 36cm 0 0},clip,width=0.245\textwidth]{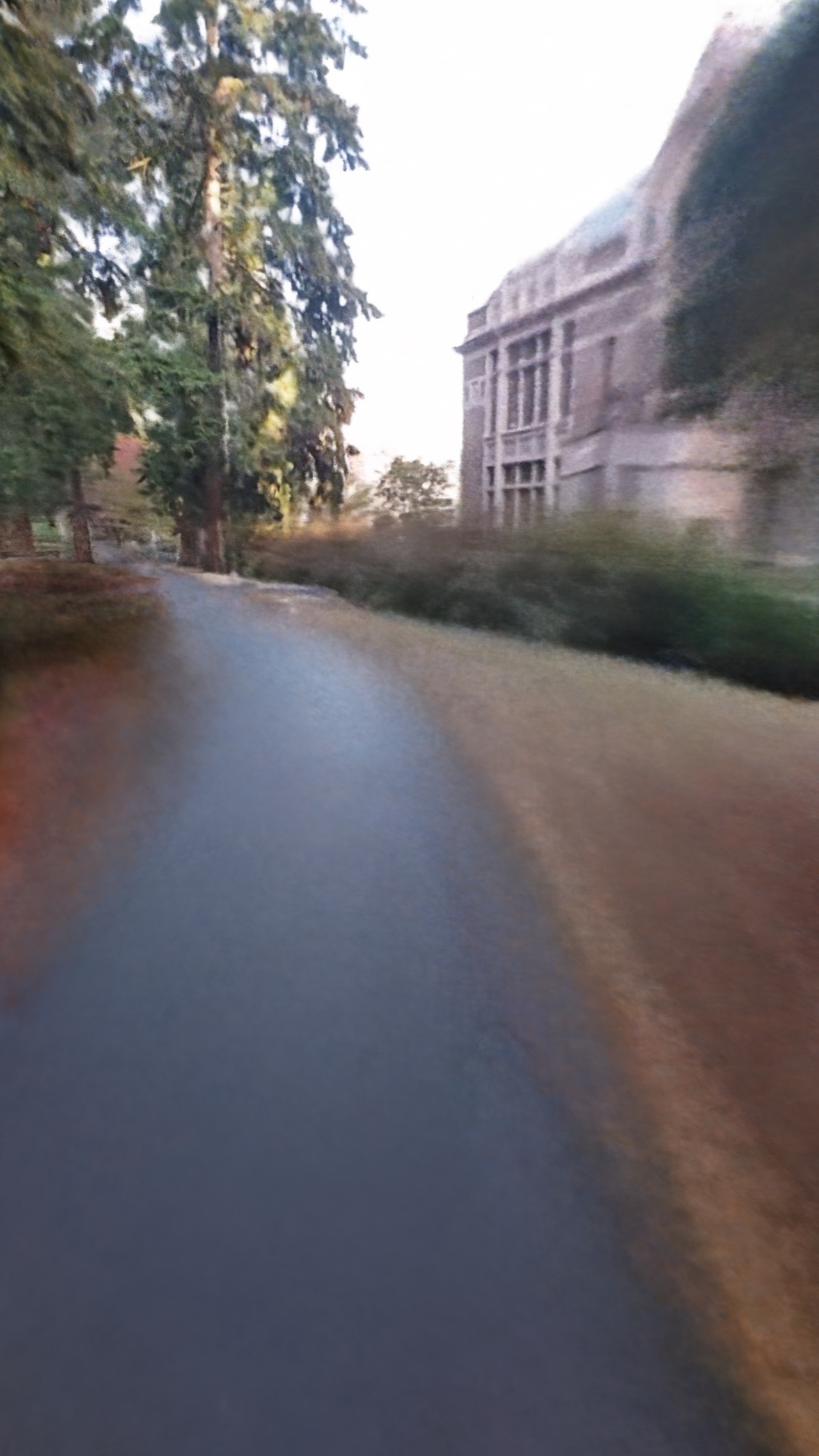}} &
\centeredtab{\includegraphics[trim={0 36cm 0 0},clip,width=0.245\textwidth]{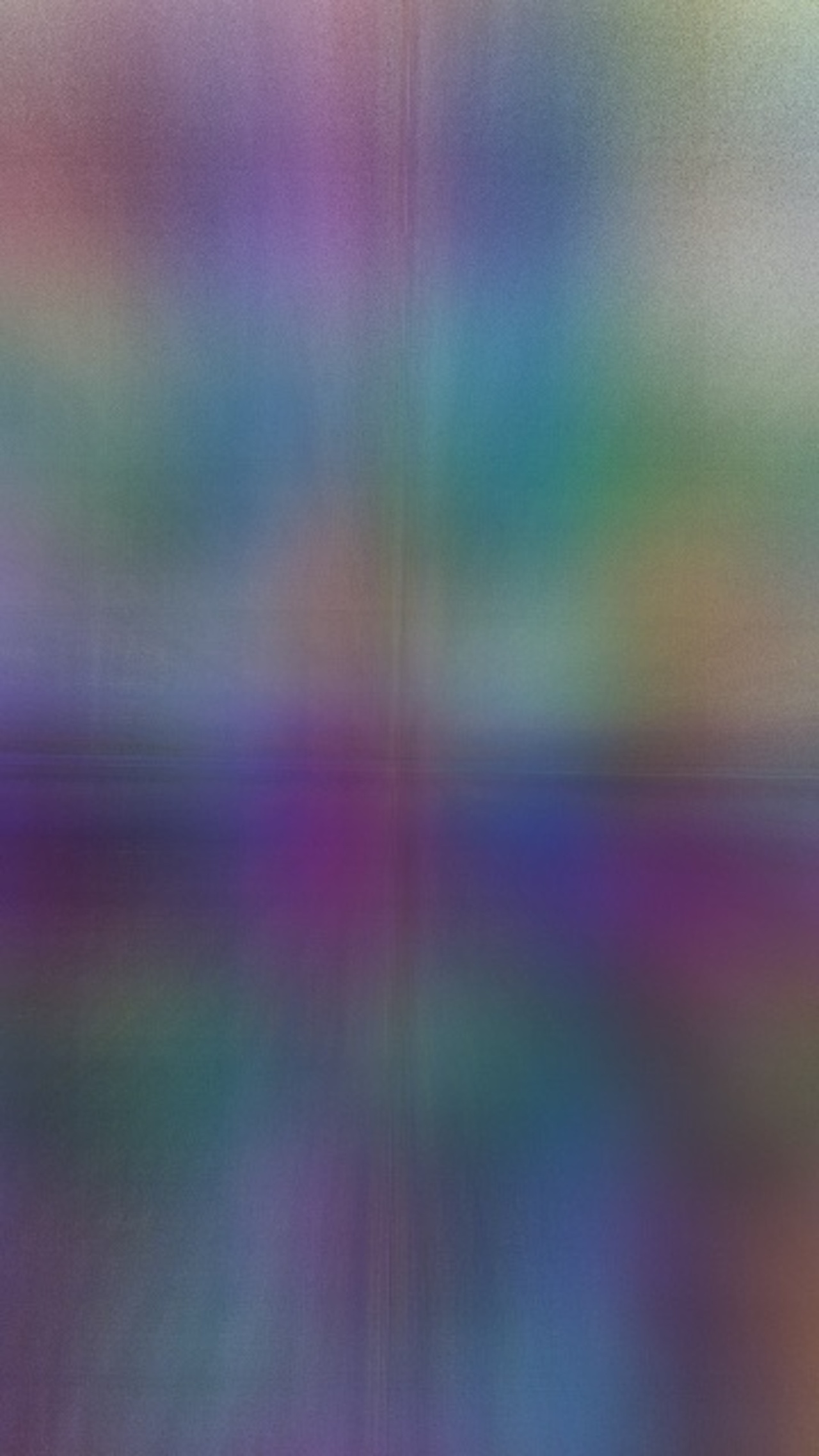}} &
\centeredtab{\includegraphics[trim={0 36cm 0 0},clip,width=0.245\textwidth]{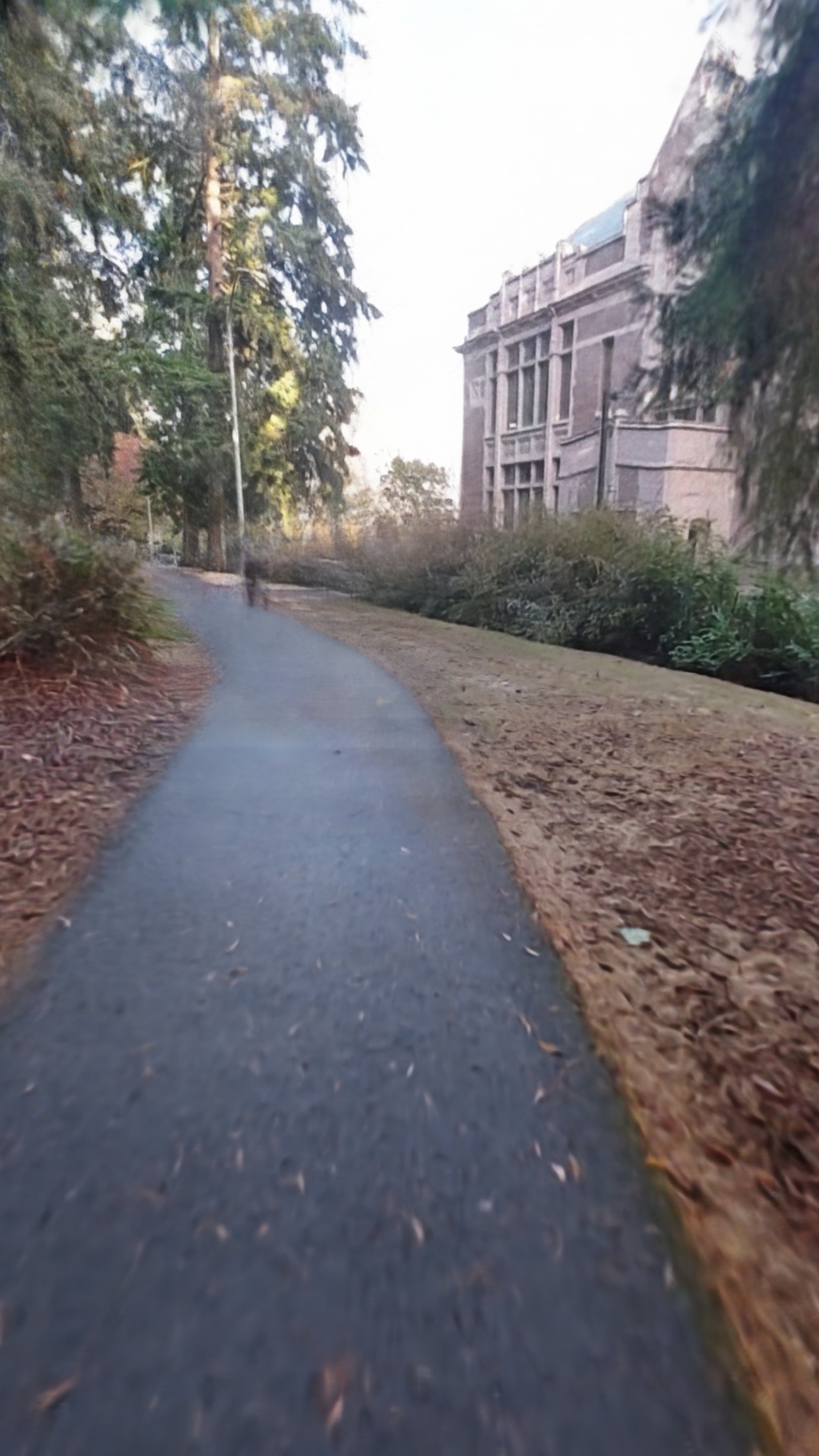}}
\\
Ground truth & Mip-NeRF360 & BARF & LocalRF (ours) \\
\end{tabular}%

\vspace{-3mm}
\caption{\textbf{Novel view synthesis results on the \textsc{Static Hikes} dataset.} Local radiance fields allow us to maintain sharpness throughout the trajectory. Some Mip-NeRF360 results, relying on preprocessed poses, are missing. Our method can optimize poses robustly, which allows good results even in scenes where other methods are less reliable.
}
\vspace{-3mm}
\label{fig:ours_results}
\end{figure*}

%% file: figure/extrapolation_main.tex
\begin{figure}[ht]
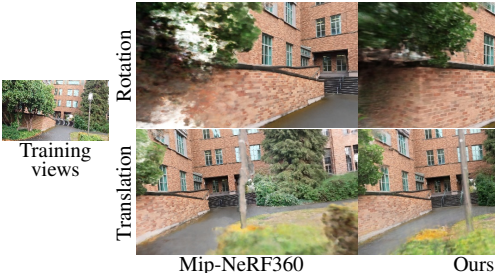

\centering
\footnotesize
\setlength{\tabcolsep}{1pt}
\renewcommand{\arraystretch}{0.3}
\begin{tabular}{cc}
\centeredtab{\animategraphics[width=40px]{30}{fig/extra/in3/}{0}{49} \\ Training \\ views} & 
\begin{tabular}{cc}
\centeredtab{\rotatebox[origin=c]{90}{Rotation}} & 
\centeredtab{\animategraphics[loop,height=47px]{10}{fig/extra/r3/}{0}{15}} \\
\centeredtab{\rotatebox[origin=c]{90}{Translation}} &
\centeredtab{\animategraphics[loop,height=47px]{10}{fig/extra/t3/}{0}{39}} \\
\end{tabular}
\\
 & \hspace{8mm} Mip-NeRF360 \hfill Ours \hspace{11mm} 
\end{tabular}%
\vspace{-3mm}
\caption{\small\textbf{Input path deviation.} We can render novel views that deviate from input path.
Please use Adobe Reader and click on images to see the embedded \emph{animations}.
}
\label{fig:extra}
\vspace{-1.5mm}
\end{figure}

%% file: table/ablation.tex
\begin{table}[t]
    \footnotesize
    \caption{
    \textbf{Ablation study.}
    We report PSNR, SSIM and LPIPS on five scenes of the \textsc{Static Hikes} dataset. 
    }
    \vspace{-3mm}
    \label{tab:ablation}
    \centering
    \begin{tabular}{l | ccc}
    \toprule
    & PSNR $\uparrow$ & SSIM $\uparrow$ & LPIPS $\downarrow$ \\
    \midrule
    Ours w/o progressive optimization   & 14.55 & 0.275 & 0.918 \\
    Ours w/o local RF                   & 16.00 & 0.306 & 0.838 \\
    LocalRF (ours)                                & \textbf{18.83} & \textbf{0.507} & \textbf{0.564} \\
    \bottomrule
    \end{tabular}
    \vspace{-3mm}
\end{table}

%% file: 5_limitations.tex
\subsection{Limitations}
\label{sec:limitations}
\noindent
We show that our method can estimate long camera trajectories robustly while maintaining a high-resolution representation. 
However, our pose estimation and progressive scheme assume that we are working with a continuous video \emph{without shot changes}.
This means that our method is not fit to reconstruct a scene from a collection of unstructured frames without coherence. 
We also do not tackle dynamic elements. 
Dynamic elements in the last row of Figure~\ref{fig:ours_results} lead to blurry regions. 
Another limitation that we observed is that sudden rotations can break pose estimation, leading to poorly rendered images.

%% file: 6_conclusions.tex
\section{Conclusions}
\label{sec:conclusions}
\noindent
We have presented a new method for reconstructing radiance fields of a large scene from a casually captured video. 
The core ideas of our work are 
1) a \emph{progressive} optimization scheme for jointly estimating camera poses and radiance fields and 
2) dynamically instantiating \emph{local} radiance fields.
From extensive evaluation on two datasets, we show that the proposed method substantially improves the robustness and fidelity of radiance field reconstruction in this challenging scenario. 